\documentclass{article}

\usepackage{microtype}
\usepackage{graphicx}
\usepackage{subfigure}
\usepackage{booktabs} 

\usepackage{hyperref}

\usepackage[accepted]{icml2024}

\usepackage{amsmath}
\usepackage{amssymb}
\usepackage{mathtools}
\usepackage{amsthm}

\theoremstyle{plain}
\newtheorem{theorem}{Theorem}[section]

\theoremstyle{definition}
\newtheorem{definition}[theorem]{Definition}

\theoremstyle{remark}

\theoremstyle{example}
\newtheorem{example}[theorem]{Example}

\usepackage{textgreek} 

\usepackage{array}
\usepackage{stfloats}
\usepackage{threeparttable}
\usepackage{multirow}
\usepackage[textsize=tiny]{todonotes}

\icmltitlerunning{A parameter-free clustering algorithm for missing datasets}

\begin{document}

\twocolumn[
\icmltitle{A parameter-free clustering algorithm for missing datasets}



\icmlsetsymbol{equal}{$\dag$}

\begin{icmlauthorlist}
\icmlauthor{Qi Li}{equal,sch}
\icmlauthor{Xianjun Zeng}{equal,sch}
\icmlauthor{Shuliang Wang \textsuperscript{*}}{sch}
\icmlauthor{Wenhao Zhu}{sch}
\icmlauthor{Shijie Ruan}{sch}
\icmlauthor{Zhimeng Yuan}{sch}
\end{icmlauthorlist}

\icmlaffiliation{sch}{School of Computer Science \& Technology, Beijing Institute of Technology, Beijing, China. Email: Qi Li $<$liqi\_bitss@163.com$>$, Xianjun Zeng $<$XianJTeng@163.com$>$, Shuliang Wang $<$slwang2011@bit.edu.cn$>$, Wenhao Zhu $<$1149932866@qq.coom$>$, Shijie Ruan $<$sjruan@bit.edu.cn$>$, Zhimeng Yuan $<$Yuan\_ZhiMeng@outlook.com$>$}
\icmlcorrespondingauthor{Shuliang Wang}{slwang2011@bit.edu.cn}

\icmlkeywords{Clustering, Missing data}

\vskip 0.3in
]



\printAffiliationsAndNotice{\icmlEqualContribution} 

\begin{abstract}
	Missing datasets, in which some objects have missing values in certain dimensions, are prevalent in the Real-world. Existing clustering algorithms for missing datasets first impute the missing values and then perform clustering. However, both the imputation and clustering processes require input parameters. Too many input parameters inevitably increase the difficulty of obtaining accurate clustering results. Although some studies have shown that decision graphs can replace the input parameters of clustering algorithms, current decision graphs require equivalent dimensions among objects and are therefore not suitable for missing datasets. To this end, we propose a \underline{S}ingle-\underline{D}imensional \underline{C}lustering algorithm, \emph{i.e.}, SDC. SDC, which removes the imputation process and adapts the decision graph to the missing datasets by splitting dimension and ``partition intersection" fusion, can obtain valid clustering results on the missing datasets without input parameters. Experiments demonstrate that, across three evaluation metrics, SDC outperforms baseline algorithms by at least 13.7\%(NMI), 23.8\%(ARI), and 8.1\%(Purity). 
\end{abstract}

\section{Introduction}
\label{Introduction}

	Clustering is one of important data mining techniques. It can divide objects into different clusters based on similarity, with each cluster corresponding to a specific category. Clustering has a wide range of applications, such as atmospheric prevention, image processing, bioinformatics, and so on \cite{Wang2016AutomaticCV}.  

\begin{figure}[t]
		\vskip 0.1in
		\begin{center}
		\centerline{\includegraphics[width=0.8\columnwidth]{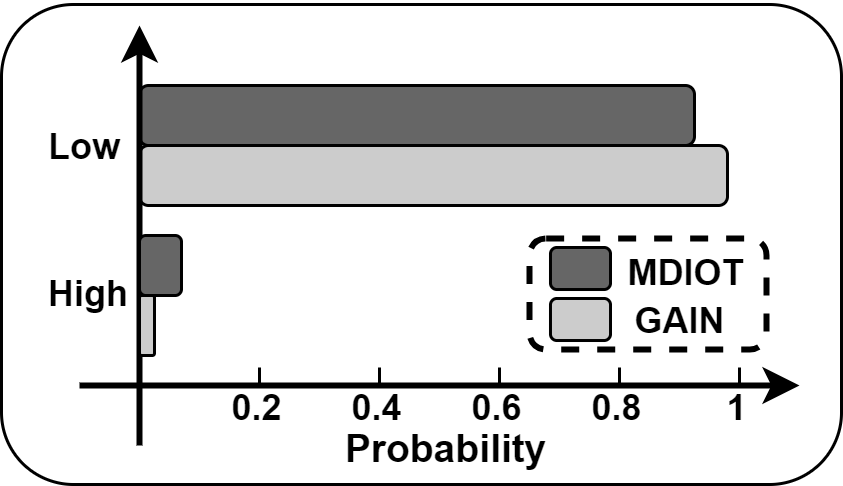}}
		\caption{The probabilities of achieving high and low accuracy for GAIN and MDIOT with different parameters.}
		\label{fig1}
		\end{center}
		\vskip -0.2in
	\end{figure}

	As clustering technology advances, there is a growing focus on enhancing clustering algorithms to handle specialized datasets \cite{Dinh2021ClusteringMN, CHEN2020104824}. \textbf{Missing datasets} are a representative class of specialized datasets where some objects have missing values in certain dimensions. Missing datasets are prevalent in the Real-world. Currently, a common strategy is to address missing values by imputing them, and then perform a standard clustering process \cite{KARIMZADEH2019265,9689944}. However, both the imputation and clustering processes require input parameters, leading to an overabundance of input parameters for existing clustering algorithms for missing datasets. Too many input parameters inevitably increase the difficulty of obtaining accurate clustering results \cite{Wang2019RobustCW}. Figure \ref{fig1} illustrates the probabilities of obtaining high and low accuracy for 2 existing algorithms with more than 3 input parameters under 100 sets of parameter values, where 0.5 is used as the threshold to distinguish high and low accuracy. The results indicate that these algorithms only have a small probability of obtaining high accuracy. In other words, a large number of parameter values fail to achieve high accuracy. Clearly, eliminating input parameters in the clustering task with missing values is a meaningful research direction.

	In recent years, some researchers have explored the use of decision graphs to replace input parameters \cite{doi:10.1126/science.1242072,LIU2018200}. They utilize decision graphs to visualize the implicit information within datasets and assist users in obtaining effective results in scenarios with either no parameters (such as DPC \cite{doi:10.1126/science.1242072}) or few parameters (such as TGP \cite{Wang2019RobustCW}). For instance, DPC generates a decision graph with density metric $\rho$ and minimum distance metric $\delta$ as its horizontal and vertical coordinates. From the decision graph of DPC, the user can observe how many objects have significantly large $\rho$ and $\delta$ to identify them as clustering centers, thus eliminating the need to set the input parameter $k$ to determine the number of clustering centers as in K-means \cite{MacQueen1967SomeMF}. However, the metrics in existing decision graphs rely on equivalent dimensions among objects, making them unsuitable for the clustering task with missing values. For example, DPC requires calculating the Euclidean distance between objects to determine the metric $\delta$. Since the Euclidean distance compares the differences in values along each dimension between objects, if a value is missing for an object in a particular dimension, the Euclidean distance cannot be calculated. \textbf{Consequently, it is a challenge to adapt decision graph to the clustering task with missing values.}
	
	In this paper, we propose a novel clustering algorithm called SDC (\underline{S}ingle-\underline{D}imensional \underline{C}lustering) to adapt decision graph to the clustering task with missing values, enabling users to obtain effective clustering results without parameters. The contributions of this work are presented as follows:
	
	\begin{itemize}
		\item We propose the first parameter-free algorithm for the clustering task with missing values. (Section \ref{principle})
		\item We devise a single-dimensional strategy for SDC to remove the imputation process and adapt the decision graph to the missing datasets. This strategy involves dividing the missing dataset into several ``non-missing" single-dimensional datasets and then obtaining clustering results for each ``non-missing" single-dimensional dataset using a density distribution decision graph. Finally, we use the proposed ``partition intersection" method to confuse these clustering results to obtain the final result. (Section \ref{clustering-process})
		\item We introduce ``gravity" to contract the boundaries of clusters, allowing the single-dimensional datasets to inherit the information from the original dataset as much as possible, thereby enhancing the effectiveness of the single-dimensional strategy. (Section \ref{cluster-information})
		\item We design a lightweight batch-density calculation method to accelerate decision graph generation and gravity calculation, significantly reducing the time-complexity of SDC. (Section \ref{lightweight-density})
		\item Extensive experiments demonstrate that, across three evaluation metrics, SDC without parameters outperforms baseline algorithms with multiple parameters by at least 13.7\%(NMI), 23.8\%(ARI), and 8.1\%(Purity). Furthermore, the advantage of SDC over baseline algorithms remains consistent regardless of the increase in missing data rate. (Section \ref{experiment})
	\end{itemize}

\section{The proposed method: SDC} 
	\label{principle}
	We introduce SDC from three aspects: the clustering process, cluster-information enhancement process, and lightweight density calculation process.
	
	\subsection{Clustering process}
	\label{clustering-process}
	\textbf{Problem Definition. } Given a dataset $X$ containing $N$ objects, $X = \{ x_{1},x_{2},\cdots,x_{N} \} \in R^{d}$, where $d$ is the dimension of the dataset. In $X$, there exist some \textbf{missing objects} with values missing in certain dimensions. $X$ is split into $d$ individual single-dimensional datasets, and the $i$-th single-dimensional dataset is denoted as $X^i$, $X^{i} = \{ y_{1},y_{2},\cdots,y_{N(i)} \}$ ($X^{i}$ is the non-missing dataset) and $N(i) \leq N$. For example, $X = \left\{ \left\langle 1,2 \right\rangle,\left\langle 5, \right\rangle,\left\langle 7,9 \right\rangle \right\}$, in which the second-dimension value is missing in the second object, then $X^{1} = \left\{ \left\langle 1 \right\rangle,\left\langle 5 \right\rangle,\left\langle 7 \right\rangle \right\}$ and $X^{2} = \left\{ \left\langle 2 \right\rangle,\left\langle 9 \right\rangle \right\}$. SDC first independently performs clustering on each single-dimensional dataset $X^{i}$ to obtain rough cluster-partition $\{ {clu(i)}_{1},{clu(i)}_{2},\cdots,{clu(i)}_{S(i)} \}$, and then fuses these rough cluster-partitions to identify the true clusters in $X$, denoted as $\left\{ {clu}_{1},{clu}_{2},\cdots,{clu}_{S} \right\}$. 
	
	\begin{figure}[t]
		\vskip 0.1in
		\begin{center}
		\centerline{\includegraphics[width=0.9\columnwidth]{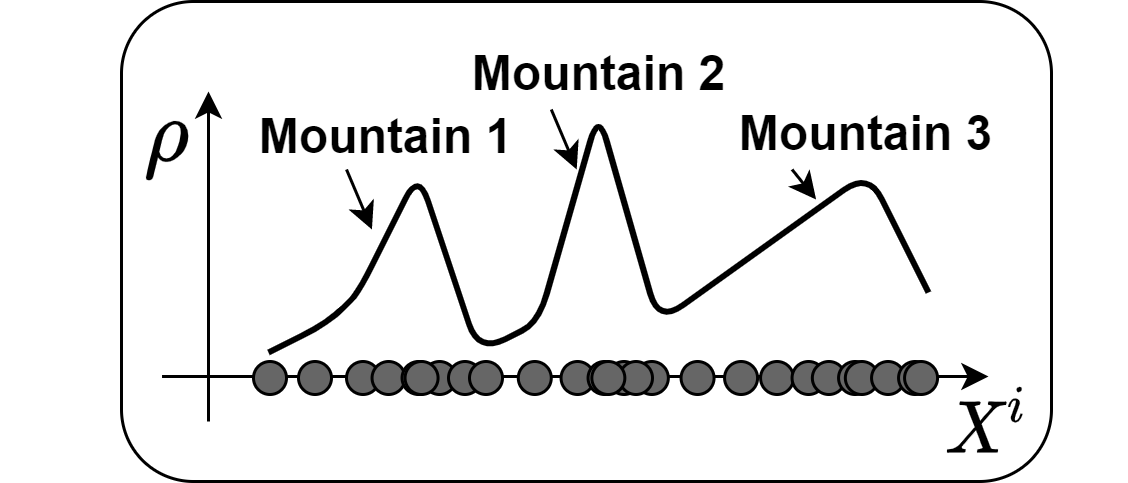}}
		\caption{An example of a density distribution decision graph}
		\label{fig2}
		\end{center}
		\vskip -0.2in
	\end{figure}
	
	\textbf{Clustering on Single-Dimensional Datasets.} For $\forall X^{i}$, as $X^{i}$ is single-dimensional, the clusters in $X^{i}$ can only be linearly arranged in the feature space of $X^{i}$, forming multiple alternating regions of high and low density. To identify all clusters in $X^{i}$, it is sufficient to recognize the high-density regions in the feature space of $X^{i}$. Hence, we calculate the density $\rho(y_{j})$ of each object $y_{j}$ in $X^{i}$ (see Section \ref{lightweight-density} for details of the density calculation). Subsequently, using the feature value of $X^{i}$ as horizontal coordinate and density $\rho$ as vertical coordinate, we generate a density distribution plot for $X^{i}$, which we refer to as the decision graph, as shown in Figure \ref{fig2}. From this decision graph, without any input parameters, we can easily distinguish density mountains by naked eyes, with each mountain corresponding to a high-density region, \emph{i.e.}, a cluster. Let us assume that $X^{i}$ is partitioned into $S(i)$ clusters, and we denote this cluster-partition as $div(i) = \{ {clu(i)}_{1},{clu(i)}_{2},\cdots,{clu(i)}_{S(i)} \}$. Since the objects in $X^{i}$ correspond to the objects in $X$, $div(i)$ can also be viewed as a cluster-partition for $X$. We proved in Theorem \ref{theorem-1} that objects that do not belong to the same cluster in $div(i)$ must not belong to the same true cluster. Hence, the cluster-partition $div(i)$ obtained on single-dimensional dataset $X^{i}$ is useful for identifying true clusters in $X$. The proofs for all the theorems in this paper are provided in Appendix \ref{appendix_theorems}.
	
	\begin{theorem}
		Let $div(i)$ be $\{ {clu(i)}_{1},{clu(i)}_{2},\cdots,{clu(i)}_{S(i)}\}$ and $\{{clu}_{1},{clu}_{2},\cdots,{clu}_{S} \}$ be the true clusters in $X$. For $\forall x_{g} \in {clu(i)}_{k}$ and $\forall x_{r} \in {clu(i)}_{l}$, if ${clu(i)}_{k} \neq {clu(i)}_{l}$, then $\nexists{clu}_{p} \in \left\{ {clu}_{1},{clu}_{2},\cdots,{clu}_{S} \right\}$, such that $x_{g},x_{r} \in {clu}_{p}$.
		\label{theorem-1}
	\end{theorem}
	
	\textbf{Fusion: Partition Intersection.} Obviously, the true clusters in $X$ must be determined jointly by multiple dimensions of $X$. To accurately identify the true clusters, it is necessary to fuse the cluster-partitions obtained on single-dimensional datasets. Specifically, for the cluster-partitions $div(i)$ and $div(i - 1)$ obtained on $X^{i}$ and $X^{i - 1}$ respectively, we in Definition \ref{def-1} define the ``partition intersection" to fuse them.
	
	\begin{definition}
		\textbf{(Partition intersection)} The partition intersection between $div(i)$ and $div(i - 1)$ is denoted	as $div(i) \cap div(i - 1)$, $div(i) \cap div(i - 1) = \big\{ {clu(i)}_{k} \cap {clu(i - 1)}_{l} \big| \forall{clu(i)}_{k} \in div(i)$, $\forall{clu(i - 1)}_{l} \in div(i - 1) \big\}$, in which ${clu(i)}_{k} \cap {clu(i - 1)}_{l}$ is called fusion cluster.
		\label{def-1}
	\end{definition}
	
	\begin{example}
		\textbf{(Partition intersection)} $div(i) = \left\{ \left\{ x_{1},x_{2},x_{4},x_{6} \right\},\left\{ x_{3},x_{5},x_{7} \right\} \right\}$, $div(i - 1) = \left\{ \left\{ x_{1},x_{2} \right\},\left\{ x_{3},x_{4},x_{5},x_{6},x_{7} \right\} \right\}$,	then $div(i) \cap div(i - 1) = \left\{ \left\{ x_{1},x_{2} \right\},\left\{ x_{3},x_{5},x_{7} \right\},\left\{ x_{4},x_{6} \right\} \right\}$. $\left\{ x_{1},x_{2} \right\}$ is one of fusion clusters.
		\label{ex-1}
	\end{example}
	
	We proved in Theorem \ref{theorem-2} that objects that do not belong to the same fusion cluster must not belong to the same true cluster. Hence, it is effective to utilize the partition intersection to fuse cluster-partitions.
	
	\begin{theorem}
		$div(i) \cap div(i - 1) = \big\{ {clu(i)}_{k} \cap {clu(i - 1)}_{l} \big| \forall{clu(i)}_{k} \in div(i),\ \forall{clu(i - 1)}_{l} \in div(i - 1) \big\}$ is the partition intersection between $div(i)$ and $div(i - 1)$. $\left\{{clu}_{1},{clu}_{2},\cdots,{clu}_{S} \right\}$ is the true cluster in $X$. For $\forall x_{g} \in {clu(i)}_{k} \cap {clu(i - 1)}_{l}$, $\forall x_{r} \in {clu(i)}_{p} \cap {clu(i - 1)}_{q}$, if ${clu(i)}_{k} \cap {clu(i - 1)}_{l} \neq {clu(i)}_{p} \cap {clu(i - 1)}_{q}$, then $\nexists{clu}_{f} \in \left\{ {clu}_{1},{clu}_{2},\cdots,{clu}_{S} \right\}$, such that $x_{g},x_{r} \in {clu}_{f}$.
		\label{theorem-2}
	\end{theorem}
	
	\textbf{Fusion: Missing Objects Merge.} However, since $X$ is a missing dataset, there exist some missing objects that appear only in $div(i)$ or $div(i - 1)$. According to Definition \ref{def-1}, the partition intersection can not assign each missing object to a certain fusion cluster, which is unreasonable. Therefore, after the partition intersection, we merge missing objects into the fusion clusters where the non-missing objects are located. We illustrate this merging process with an example in Example \ref{ex-2}.
	
	\begin{example}
		\textbf{(Missing objects merge)} $div(i) = \left\{ \left\{ x_{1},x_{2},x_{4},x_{6} \right\},\left\{ x_{3},x_{5},x_{7} \right\} \right\}$, $div(i - 1) = \left\{ \left\{ x_{1},x_{2} \right\},\left\{ x_{3},x_{4}{,x}_{5},x_{6},x_{7},x_{8} \right\} \right\}$. The value of $x_{8}$ is missing in the $i$-th dimension, so it does not appear in $div(i)$. According to Theorem \ref{theorem-1}, $x_{8}$ can only belong to the same true cluster as $x_{3},x_{4},x_{5},x_{6}$ or $x_{7}$. However, the partition intersection $div(i) \cap div(i - 1)$ assigns $x_{3},x_{4},x_{5},x_{6},x_{7}$ into two fusion clusters, namely $\left\{ x_{3},x_{5},x_{7} \right\}$ and $\left\{ x_{4},x_{6} \right\}$. Since $\left\{ x_{3},x_{5},x_{7} \right\}$ contains more objects, the probability of $x_{8}$ belonging to the same true cluster as $\left\{ x_{3},x_{5},x_{7} \right\}$ is higher. Therefore, we merge $x_{8}$ into the fusion cluster containing $\left\{ x_{3},x_{5},x_{7} \right\}$.
		\label{ex-2}
	\end{example}
	
	After fusing all cluster-partitions, the final fusion clusters are the clusters identified by SDC from $X$. Algorithm \ref{alg:algorithm1} describes the detailed implementation of the clustering process.
	
	\begin{algorithm}[h]
		\caption{Clustering progress}
		\label{alg:algorithm1}
		\begin{algorithmic}[1]
		\STATE{\bfseries Input:} Dataset $X$
		\STATE{\bfseries Output:} Clustering result $div(d)$
		\FOR{$i=1$ {\bfseries to} $d$}
			\STATE Calculate density of objects in $X^i$ and draw decision graph.
			\STATE Obtain $div(i)$ according to the decision graph.
			\IF{$i>1$} 
				\STATE Fuse $div(i)$ and $div(i-1)$ according to Definition \ref{def-1}.
				\STATE Merge missing objects into existing fusion clusters according to Example \ref{ex-2}.
				\STATE $div(i) \leftarrow$ fusion result.
			\ENDIF
		\ENDFOR
		\end{algorithmic}
	\end{algorithm}
	
	\begin{figure}[h]
		\vskip 0.1in
		\begin{center}
			\includegraphics[width=1.2\columnwidth]{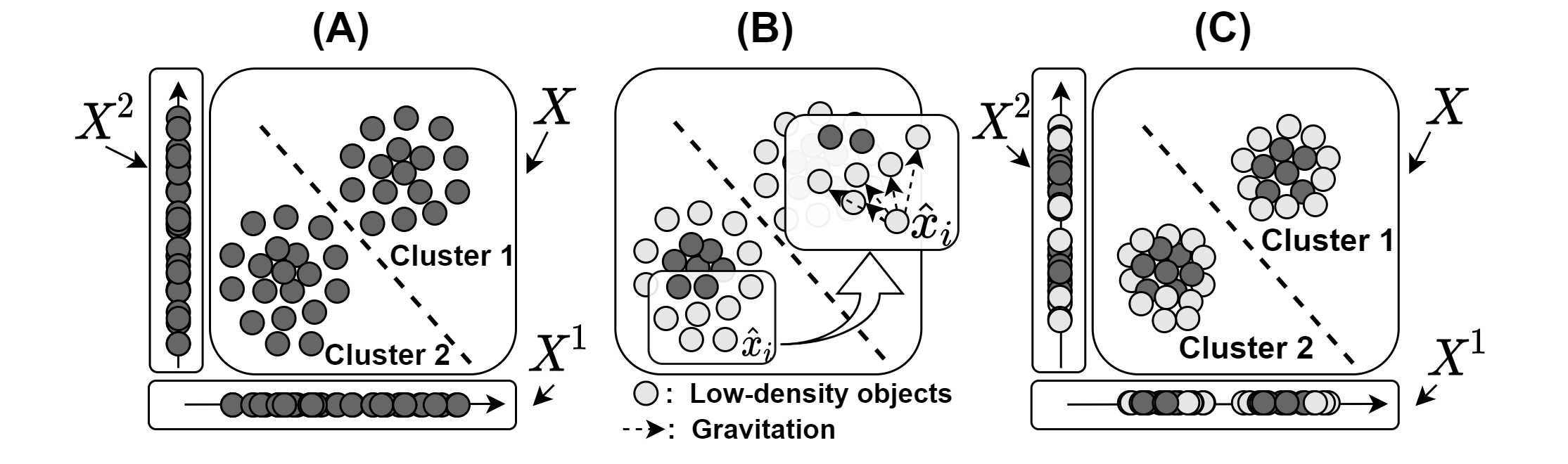}
			\caption{An example of the cluster-information enhancement process.}
			\label{fig3}
		\end{center}
		\vskip -0.1in
	\end{figure}
	
	\subsection{Cluster-information enhancement process}
	\label{cluster-information}
	\textbf{Motivation.} The more cluster-information of $X$ is retained in $X^{i}$, the higher the accuracy of the clustering process is. However, in some scenarios, $X^{i}$ may lose the cluster-information of $X$. As shown in Figure \ref{fig3} (A), for a dataset containing two linearly separable clusters, its $X^{1}$ and $X^{2}$ have only one cluster. The reason for this is that the clusters in $X$ are close to each other. Therefore, before splitting $X$ into $d$ single-dimensional datasets, it is necessary to contract the cluster boundaries in $X$ (\emph{i.e.}, increase the inter-cluster distance) to enhance the cluster-information of the single-dimensional datasets. Since the density of objects at the cluster boundaries is usually lower than that inside the clusters \cite{doi:10.1126/science.1242072}, the cluster boundaries can be contracted simply by forcing neighboring low-density objects in $X$ to be closer to each other. According to Section \ref{clustering-process}, identifying clusters in $X$ relies heavily on non-missing objects (missing objects are assigned into the clusters where non-missing objects are located), so the cluster-information is mainly reflected by non-missing objects. Therefore, in the cluster-information enhancement process, we focus only on the non-missing objects in $X$.	
	
	\textbf{Gravitational Calculation.} Let $\widehat{X}$ be the set of non-missing objects in $X$, denoted as $\widehat{X} = \left\{ {\widehat{x}}_{1},{\widehat{x}}_{2},\cdots,{\widehat{x}}_{M} \right\} \subset X$. Inspired by related works on gravitational clustering \cite{Li2020HIBOGIT,LI202352}, we introduce gravitational forces among objects in $\widehat{X}$, such that low-density objects (see Definition \ref{def-2} for details) are attracted to neighboring objects, forcing them to move closer to each other. Specifically, if ${\widehat{x}}_{i}$ is a low-density object, then the gravitational force we add between it and its $j$-th nearest neighbor ${\widehat{x}}_{ij}$ is
	
	\begin{equation}
		F_{ij} = \frac{G\left\| {\widehat{x}}_{i1} - {\widehat{x}}_{ij} \right\|_{2}({\widehat{x}}_{ij} - {\widehat{x}}_{i})}{\left\| {\widehat{x}}_{i} - {\widehat{x}}_{ij} \right\|_{2}^{2}},
		\label{equation1}
	\end{equation}
	where $G = \frac{1}{M}\sum_{{\widehat{x}}_{k} \in \widehat{X}}^{}\left\| {\widehat{x}}_{k1} - {\widehat{x}}_{k} \right\|_{2}$ is a constant to control the magnitude of the gravitational forces. To avoid ${\widehat{x}}_{i}$ moving toward distant objects, we stipulate that only the five nearest neighbors $\left\{ {\widehat{x}}_{i1},{\widehat{x}}_{i2},{\widehat{x}}_{i3},{\widehat{x}}_{i4},{\widehat{x}}_{i5} \right\}$ apply gravitational forces on ${\widehat{x}}_{i}$, as shown in Figure \ref{fig3} (B). Therefore, the total force applied on ${\widehat{x}}_{i}$ is $F_{i} = \sum_{j = 1}^{5}F_{ij}$.
	
	\begin{definition}
		\textbf{(Low-density object)} The average density of objects in $\widehat{X}$ is $\overline{\rho} = \frac{1}{M}\sum_{i = 1}^{M}{\rho\left( {\widehat{x}}_{i} \right)}$, where $\rho({\widehat{x}}_{i})$ is the density of ${\widehat{x}}_{i}$ in $\widehat{X}$ (the density calculation process is detailed in Section \ref{lightweight-density}). For $\forall{\widehat{x}}_{i} \in \widehat{X}$, if $\rho({\widehat{x}}_{i}) < \overline{\rho}$, we refer to $\ {\widehat{x}}_{i}$ as a low-density object.
		\label{def-2}
	\end{definition}
	
	\textbf{Object Movement.} We adopt the uniform acceleration motion model ($S = \frac{1}{2} \cdot \frac{F}{m}t^{2} + v_{0}t$, where $F$ is the applied force, $m$ is the object's mass, $t$ is the duration, and $v_{0}$ is the initial velocity) to calculate the displacement of ${\widehat{x}}_{i}$ under $F_{i}$. 
	Since data object motion is not really a physical event, mass($m$), time($t$) and velocity($v_0$) are virtual and meaningless. So we set $m = 1$, $t = 1$, and $v_0 = 0$ to eliminate them ($m$ and $t$ are set to 1 instead of 0 because they are coefficients of other variables). Eventually, the motion model for ${\widehat{x}}_{i}$ becomes $S_{i} = \frac{1}{2}F_{i}$. After movement, ${\widehat{x}}_{i}$ changes to $\boxed{{\widehat{x}}_{i}}$,
	
	\begin{equation}
		\boxed{{\widehat{x}}_{i}} = {\widehat{x}}_{i} + S_{i} = {\widehat{x}}_{i} + \frac{1}{2}\sum_{j = 1}^{5}\frac{G\left\| {\widehat{x}}_{i1} - {\widehat{x}}_{ij} \right\|_{2}({{\widehat{x}}_{ij} - \widehat{x}}_{i})}{\left\| {\widehat{x}}_{i} - {\widehat{x}}_{ij} \right\|_{2}^{2}}.
		\label{equation2}
	\end{equation}
	
	When all low-density objects in $\widehat{X}$ change according to Equation (\ref{equation2}), the distances between neighboring low-density objects will be reduced. Example \ref{ex-3} illustrates an instance of the cluster-information enhancement process, and its detailed implementation is described in Algorithm \ref{alg:algorithm2}.
	
	\begin{example}
		\textbf{(Cluster-information enhancement)} For the dataset in Figure \ref{fig3} (A), Figure \ref{fig3} (C) displays the distribution of all low-density objects after movement. Compared to the original dataset, the cluster boundaries are visibly contracted, and both $X^{1}$ and $X^{2}$ exhibit two clusters, effectively inheriting the original dataset's cluster-information.
		\label{ex-3}
	\end{example}
	
	\begin{algorithm}[h]
		\caption{Cluster-information enhancement process}
		\label{alg:algorithm2}
		\begin{algorithmic}[1]
			\STATE {\bfseries Input:} $\widehat{X}$
			\STATE {\bfseries Output:} $\boxed{\widehat{X}}$
			\FOR{$\widehat{x}_i$ {\bfseries in} $\widehat{X}$}
			\IF{$\rho(\widehat{x}_i) < \overline{\rho}$}
			\FOR{$j=1$ {\bfseries to} $5$}
			\STATE Calculate $F_{ij}$ according to Equation (1).
			\ENDFOR
			\STATE Calculate $\boxed{\widehat{x}_i}$ according to Equation (2).
			\ENDIF
			\ENDFOR
		\end{algorithmic}
	\end{algorithm}

\subsection{Lightweight Density Calculation Process}
	\label{lightweight-density}
	Before introducing density, we first provide a common definition of the neighbor-set \cite{WANG202124}.
	
	\begin{definition}
		\textbf{(Neighbor-set \cite{WANG202124})}  Given a dataset $X$, for $\forall x_{i} \in X$, its neighbor-set is $\Gamma\left( x_{i} \right) = \left\{ x_{j} \in X \middle| \left\| x_{j} - x_{i} \right\|_{2} \leq R \right\}$, in which $R$ is five times the average distance between all objects in $X$ and their closest neighbors.
		\label{def-3}
	\end{definition}
	
\begin{figure}[h]
		\vskip 0.1in
		\begin{center}
		\includegraphics[width=0.8\columnwidth]{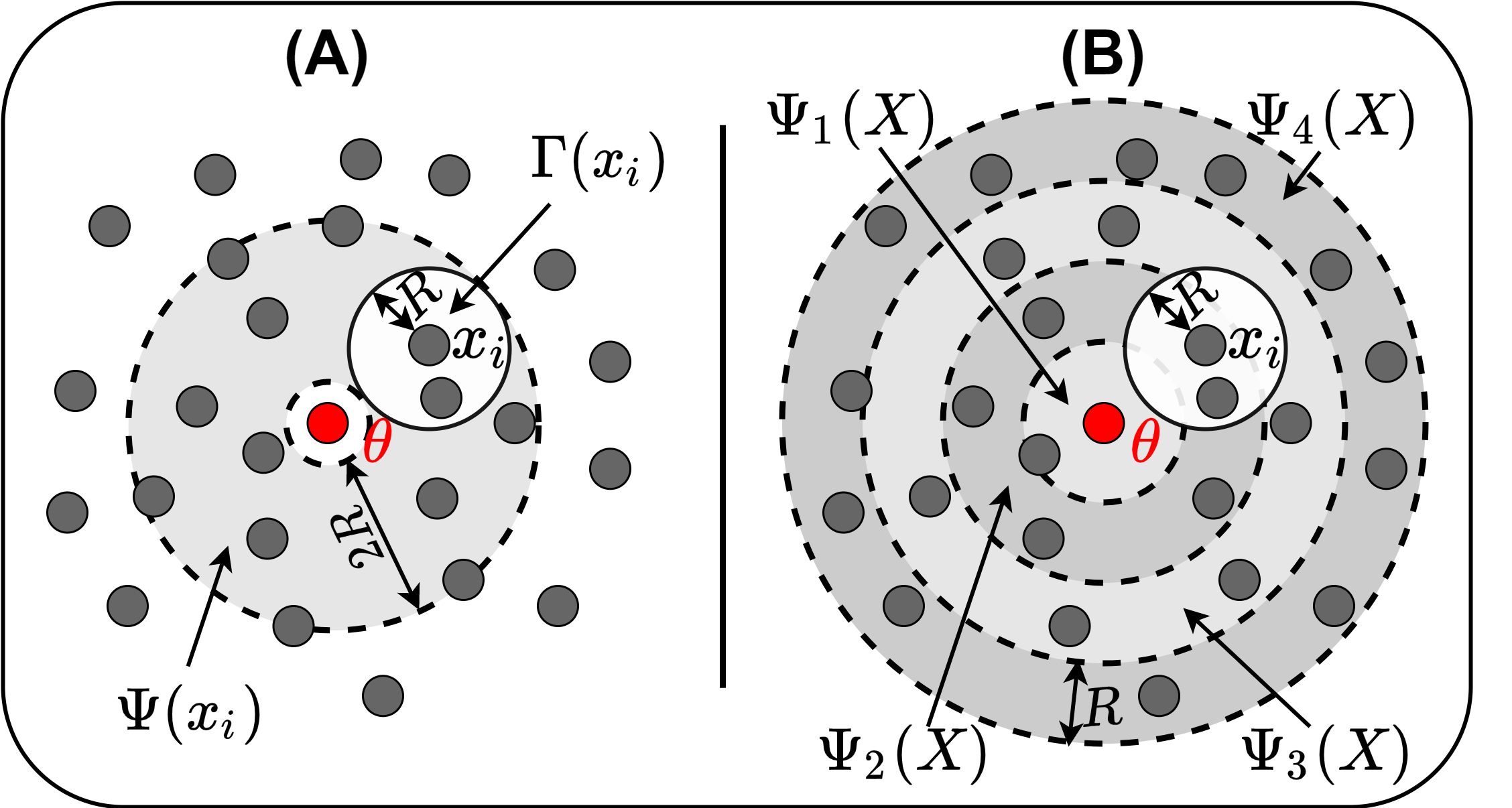}
		\caption{An example of lightweight density calculation.}
		\label{fig4}
		\end{center}
		\vskip -0.1in
	\end{figure}

	\begin{example}
		\textbf{(Neighbor-set)} As shown in Figure \ref{fig4} (A), the objects included within the circular neighborhood centered at $x_{i}$ form the neighbor-set of $x_{i}$.
		\label{ex-4}
	\end{example}
	
	\textbf{Motivation.} For $\forall x_{i} \in X$, common density calculation method \cite{doi:10.1126/science.1242072} uses the number of neighbors in the neighbor-set as the density of $x_{i}$, \emph{i.e.}, $\rho\left( x_{i} \right) = \left| \Gamma\left( x_{i} \right) \right|$. This method requires frequent distance calculations between $x_{i}$ and other objects in $X$ to determine if the other objects are within the neighbor-set of $x_{i}$. When calculating densities for all objects in $X$, its time complexity is $O(N^{2})$, which is not suitable for large-scale datasets. In this paper, we aim to eliminate unnecessary distance calculations between distant objects to significantly reduce the time complexity of density calculations in the clustering process (Section \ref{clustering-process}) and cluster-information enhancement process (Section \ref{cluster-information}). Some studies \cite{Xu2021AFD} introduced a third object to quantify the distance between two objects, leading to a lightweight adjacency matrix. Inspired by this, we introduce a virtual object as the third object to define the annular region of $x_{i}$ (see Definition \ref{def-4} and Example \ref{ex-5}), and proved in Theorem \ref{theorem-4} that $\Gamma\left( x_{i} \right)$ must be within $x_{i}$'s annular region. Therefore, the density of $x_{i}$ can be calculated by simply calculating the distance between $x_{i}$ and each object in the annular region of $x_{i}$ instead of in $X$.
	
	\begin{definition}
		\textbf{(Annular region)} For $\forall x_{i} \in X$, given a virtual object $\theta$, $\Psi\left( x_{i} \right) = \big\{ x_{j} \in X \big | \left\| x_{i} - \theta \right\|_{2} - R \leq $$ \left\| x_{j} - \theta \right\|_{2} \leq \left\| x_{i} - \theta \right\|_{2} + R \big\}$ is called the annular region of $x_{i}$.
		\label{def-4}
	\end{definition}
	
	\begin{example}
		\textbf{(Annular region)} In Figure \ref{fig4} (A), the red object is the virtual object. The gray annular region centered at the red object and containing $\Gamma(x_{i})$ is the annular region of $x_{i}$.
		\label{ex-5}
	\end{example}
	
	\begin{theorem}
		For $\forall x_{i},x_{j} \in X$, if $x_{j} \in \Gamma\left(x_{i} \right)$, then $x_{j} \in \Psi\left( x_{i} \right)$.
		\label{theorem-4}
	\end{theorem}
	
	However, when calculating the density for all objects in $X$, it is cumbersome to create a separate annular region for each object. Therefore, we divide $X$ into several public-annular regions with $\theta$ as the center, as detailed in Definition \ref{def-5}.
	
	\begin{definition}
		\textbf{(Public-annular region)} Given a virtual object $\theta$, $\Psi_{i}(X) = \left\{ x_{j} \in X \middle| (i - 1)R \leq \left\| x_{j} - \theta \right\|_{2} \leq iR \right\}$ is called the $i$-th public-annular region of $X$.
		\label{def-5}
	\end{definition}
	
	\begin{example}
		\textbf{(Public-annular region)} Figure \ref{fig4} (B) shows four public-annular regions centered at the virtual object.
		\label{ex-6}
	\end{example}
	
	We proved in Theorem \ref{theorem-5} that the neighbor-set of each object in $\Psi_{i}(X)$ must be within three adjacent public-annular regions, $\Psi_{i - 1}(X)$, $\Psi_{i}(X)$, and $\Psi_{i + 1}(X)$. Therefore, the density of all objects in $\Psi_{i}(X)$ can be ``batch" calculated by calculating the distance between objects in $\Psi_{i - 1}(X) \cup \Psi_{i}(X) \cup \Psi_{i + 1}(X)$.
	
	\begin{theorem}
		For $\forall x_{j} \in X$, if $x_{j} \in \Psi_{i}(X)$, then $\Gamma\left( x_{j} \right) \subset \Psi_{i - 1}(X) \cup \Psi_{i}(X) \cup \Psi_{i + 1}(X)$.
		\label{theorem-5}
	\end{theorem}
	
	However, there might be a large number of objects in three adjacent public-annular regions. Fortunately, we proved in Theorem \ref{theorem-6} that each object must meet the dimensionality relationship condition (see Definition \ref{def-6}) with the objects in its neighbor-set. Therefore, when ``batch" calculating the density of all objects in $\Psi_{i}(X)$, we only need to calculate the distance between objects that meet the dimensionality relationship condition in $\Psi_{i - 1}(X) \cup \Psi_{i}(X) \cup \Psi_{i + 1}(X)$. 
	
	\begin{definition}
		\textbf{(Dimensionality relationship condition)} Let $D_{i}$ and $D_{j}$ be the sums of $d$ dimensional values of $x_{i}$ and $x_{j}$, respectively. If $\left| D_{i} - D_{j} \right| \leq \sqrt{d}R$, then $x_{i}$ meets the dimensionality relationship condition with $x_{j}$.
		\label{def-6}
	\end{definition}
	
	\begin{theorem}
		For $\forall x_{i},x_{j} \in X$, if $x_{j} \in \Gamma\left( x_{i} \right)$, then $\left| D_{i} - D_{j} \right| \leq \sqrt{d}R$.
		\label{theorem-6}
	\end{theorem}

	Ultimately, the time complexity of density calculation is significantly reduced to $O(N\log N)$, as detailed in Appendix \ref{appendix_time_density}. Algorithm \ref{alg:algorithm3} describes the detailed implementation of density calculation.
	
	\begin{algorithm}[h]
		\caption{Lightweight Density Calculation Process}
		\label{alg:algorithm3}
		\begin{algorithmic}[1]
			\STATE {\bfseries Input:} $X$
			\STATE {\bfseries Output:} $\rho$
			\STATE Divide $X$ into $T$ public-annular regions according to Definition \ref{def-5}, in which $T=\lceil \max\limits_{x_j \in X}\lVert x_j - \theta \rVert_2 / R \rceil$
			\FOR{$\Psi_i(X)$ {\bfseries in} $\left\{ \Psi_1(X),\dots,\Psi_T(X) \right\}$}
			\FOR{$x_j$ {\bfseries in} $\Psi_{i-1}(X) \cup \Psi_{i}(X) \cup \Psi_{i+1}(X)$}
			\STATE Search for objects that meet the dimensionality relationship condition with $x_i$ according to Definition \ref{def-6}.
			\STATE Calculate the distance between searched objects to get $\rho(x_j)$.
			\ENDFOR
			\ENDFOR
		\end{algorithmic}
	\end{algorithm}
	
\subsection{Time Complexity Analysis} \label{time-complexity}
	
	The lightweight density calculation process is invoked by the clustering process and the cluster-information enhancement process. The complexity of the clustering process is $O(dN\log N+4dN)$; the complexity of the cluster-information enhancement process is $O(2N\log N+6N)$. Consequently, the total time complexity of SDC is $O((d+2)N\log N)$. A detailed analysis is provided in Appendix \ref{appendix_time_SDC}.

\section{Experiment}
	\label{experiment}
	\subsection{Experimental Settings} \label{settings}
	
	\textbf{Dataset.} We select some Real-world datasets from the clustering basic benchmark \cite{2018K} and some synthetic datasets from ODDS Library \cite{Rayana:2016}. Here, we adopt a popular MAR strategy \cite{liu2023improving, Tang2022DeepSI, Datta2016ClusteringWM}, which randomly removes the values of some objects in certain dimensions, to convert them to missing datasets. The detailed information of these missing datasets is recorded in Table \ref{table1}.

	\textbf{Baseline Algorithms.} We compare SDC with 10 representative clustering algorithms specialized for missing datasets, namely DIMV \cite{Vu2023ConditionalEW}, TDM \cite{10.5555/3618408.3620181}, MBFS \cite{liu2023improving}, HyperImpute \cite{pmlr-v162-jarrett22a}, MDIOT \cite{Muzellec2020MissingDI}, VAEAC \cite{ivanov2018variational}, GAIN \cite{Yoon2018GAINMD},  KPOD \cite{Chi2014kPODAM}, LRC \cite{Nie2012LowRankMR} and MICE \cite{JSSv045i03}. Since clustering is unsupervised, there are no ground-truth labels to guide the parameter selection, people can only randomly set parameters within an empirical interval. Clearly, only the average accuracy can reflect the performance of clustering algorithms in real-world. Therefore, we execute these algorithms within the recommended parameter interval and then calculate the average accuracy as their final accuracy. Their parameter intervals are provided in Appendix \ref{appendix_parameter}.

	\textbf{Evaluation Metrics.} We select three common evaluation metrics to measure clustering accuracy, namely NMI, ARI, and Purity. Specifically, ARI measures accuracy by counting the number of object pairs assigned to the same or different clusters between the experimental results and the true results. NMI measures accuracy by quantifying the normalized mutual dependence between the experimental results and the true results. Purity measures accuracy by calculating the proportion of correctly assigned objects. ARI ranges from -1 to 1, and NMI and Purity range from 0 to 1. The closer they are to 1, the more accurate the clustering results.

	\subsection{Ablation Experiments}

	We conduct ablation experiments to verify the necessity of the cluster-information enhancement process and the lightweight density calculation process for SDC.
		
	\begin{table}[ht]
		\caption{The description of datasets}
		\vskip 0.15in
		\renewcommand{\arraystretch}{0.85}
		\setlength{\tabcolsep}{3.4pt}
		\label{table1}
		\begin{center}
			\begin{small}
				\begin{tabular}{lcccr}
					\toprule
					Dataset (\emph{Abbreviation}) & Number & Dim & Cluster & Missing Rate \\
					\midrule
					\emph{\textbf{Iris}} & 150 & 4 & 3 & 20\% \\
					\emph{\textbf{Breast}} (\emph{Brea}) & 569 & 30 & 2 & 20\%\\
					\emph{\textbf{Wine}} & 178 & 13 & 3 & 60\% \\
					\emph{\textbf{Banknote}} (\emph{Bank})& 1,372 & 4 & 2 & 30\% \\
					\emph{\textbf{Htru}} & 17,898 & 8 & 2 & 60\% \\
					\emph{\textbf{Knowledge}} (\emph{Know}) & 258 & 5 & 4 & 40\% \\
					\emph{\textbf{Birch1}} (\emph{Bir1}) & 100,000 & 2 & 100 & 10\% \\
					\emph{\textbf{Birch2}} (\emph{Bir2}) & 100,000 & 2 & 100 & 30\% \\
					\emph{\textbf{Overlap1}} (\emph{Ove1})& 640 & 2 & 2 & 40\% \\
					\emph{\textbf{Overlap2}} (\emph{Ove2}) & 4,290 & 2 & 13 & 10\% \\
					\emph{\textbf{Worms}} (\emph{Worm})& 105,600 & 2 & 35 & 70\% \\
					\emph{\textbf{Urbanland}} (\emph{Urba}) & 507 & 147 & 9 & 10\% \\
					\emph{\textbf{Winnipeg}} (\emph{Winn}) & 3,258 & 174 & 7 & 50\% \\
					\bottomrule
				\end{tabular}
			\end{small}
		\end{center}
		\vskip -0.1in
	\end{table}
	
	\begin{figure}[ht]
		\vskip 0.1in
		\begin{center}
			\subfigure{
				\includegraphics[width=0.4\columnwidth]{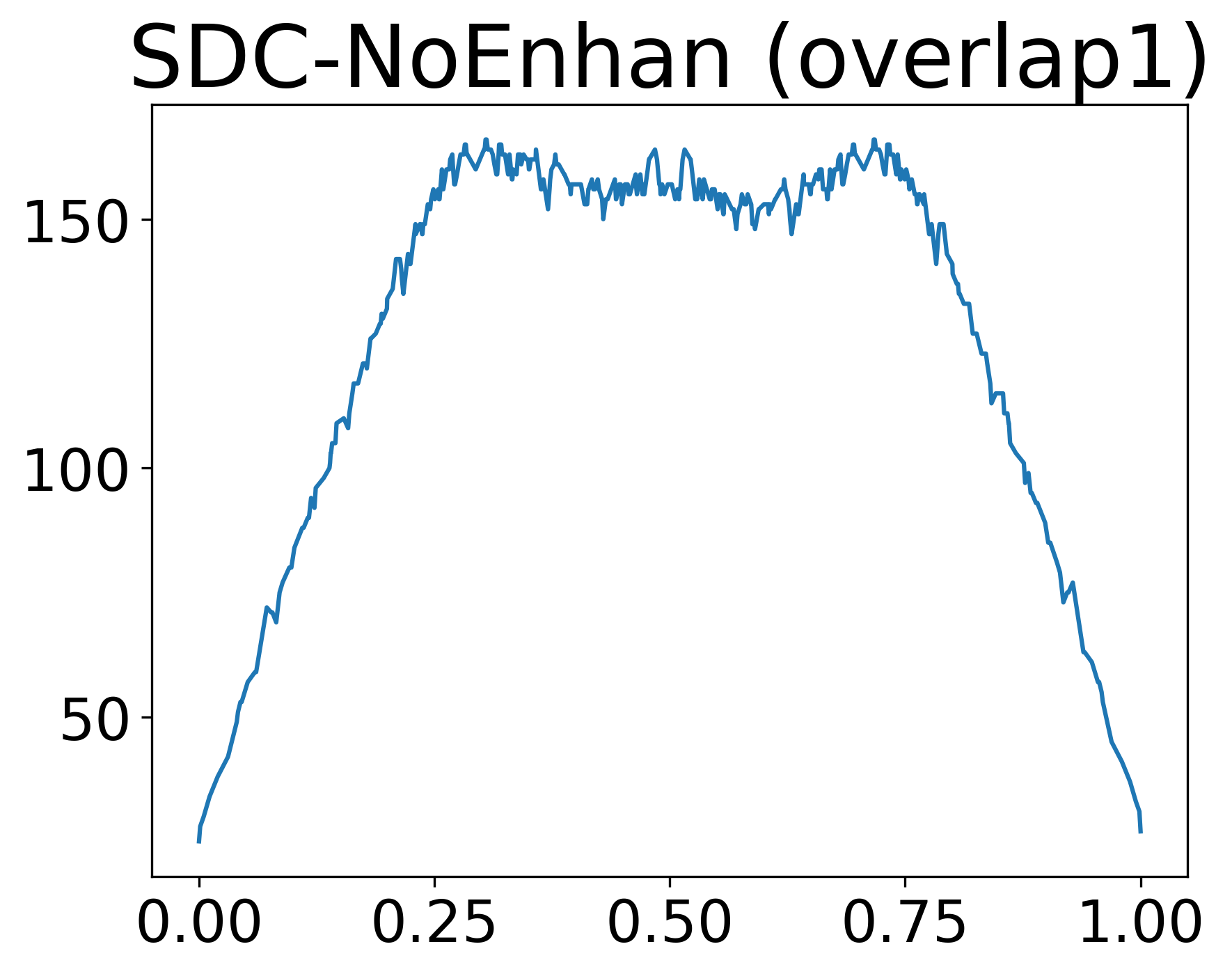}}
			\subfigure{
				\includegraphics[width=0.4\columnwidth]{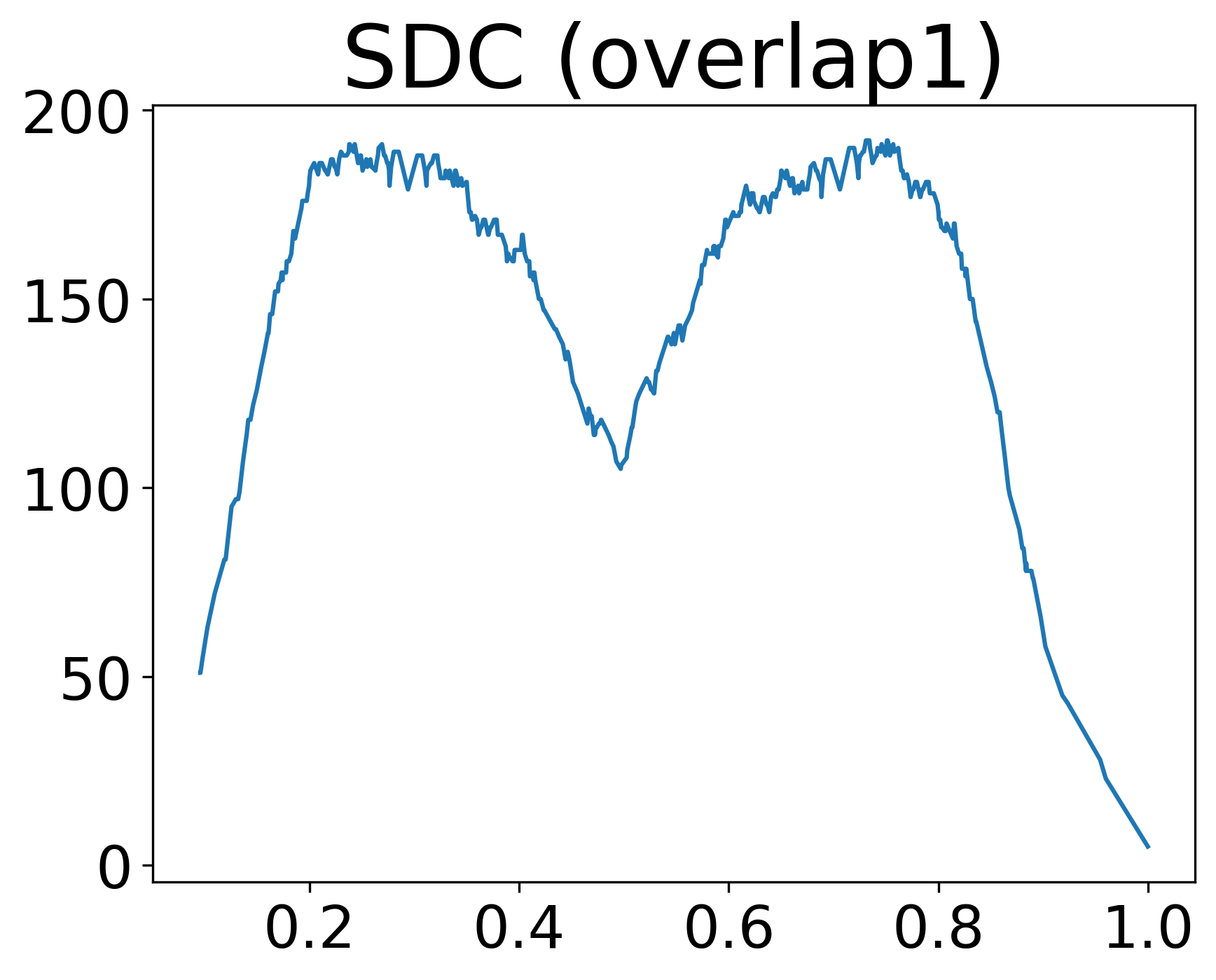}}
			\subfigure{
				\includegraphics[width=0.4\columnwidth]{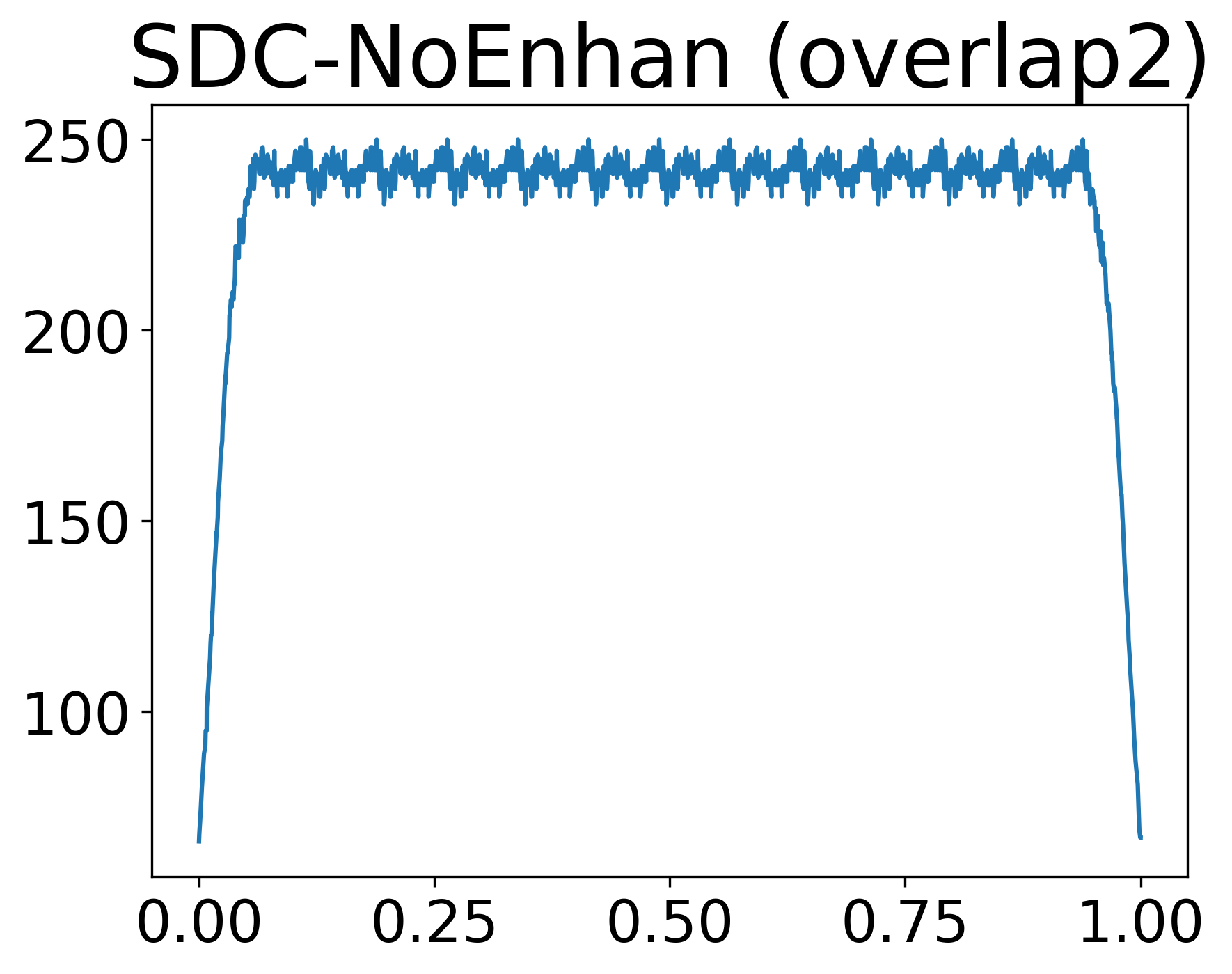}}
			\subfigure{
				\includegraphics[width=0.4\columnwidth]{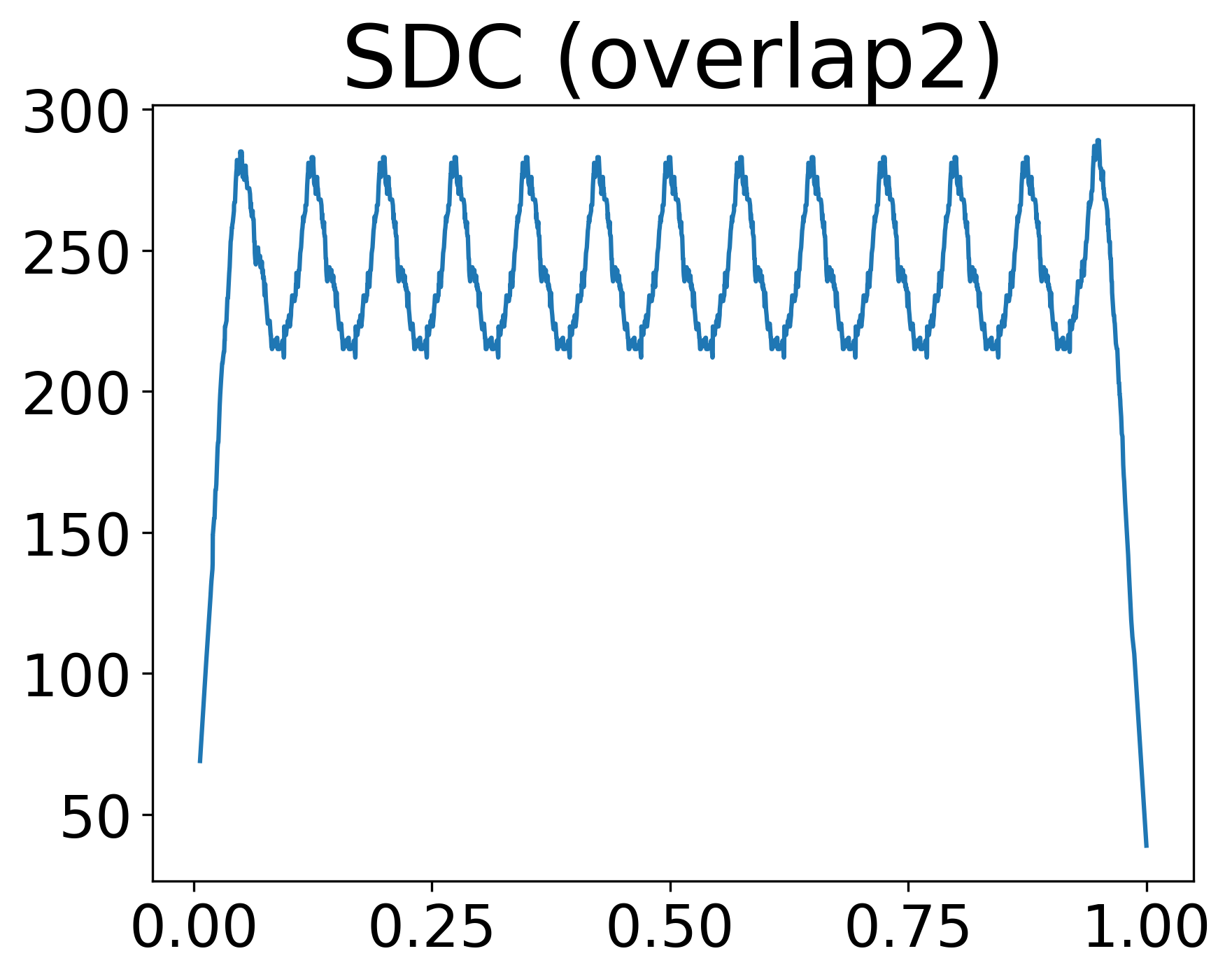}}
			\caption{The decision graphs comparison of SDC and SDC-NoEnhan on \emph{Overlap1} and \emph{Overlap2} datasets}
			\label{fig5}
		\end{center}
		\vskip -0.1in
	\end{figure}
	
	\begin{figure}[ht!]
		\vskip 0.1in
		\begin{center}
			\centerline{\includegraphics[width=0.65\columnwidth]{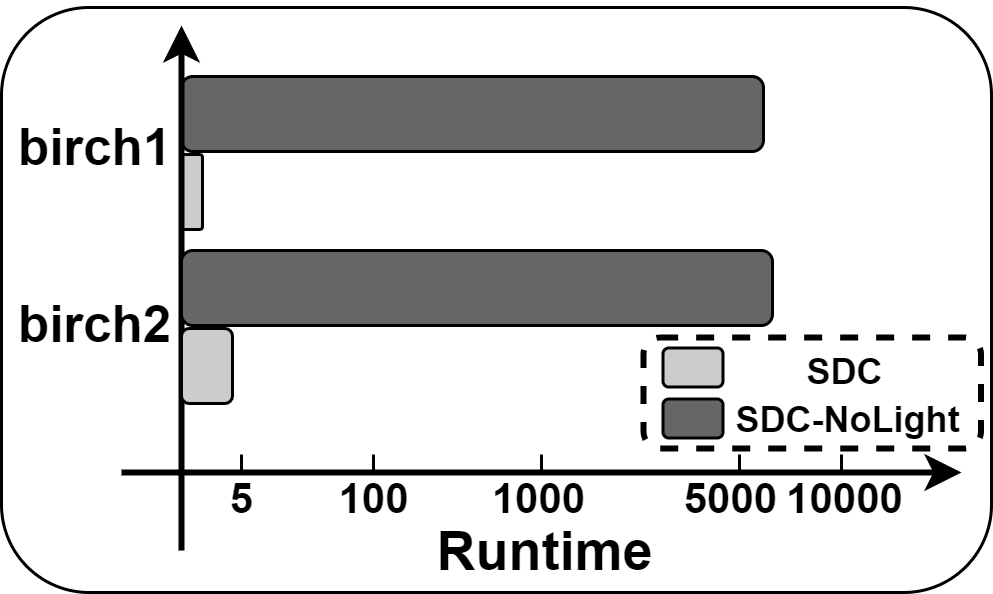}}
			\caption{The runtime comparison of SDC and SDC-NoLight on \emph{Birch1} and \emph{Birch2} datasets}
			\label{fig6}
		\end{center}
		\vskip -0.1in
	\end{figure}
	
	\textbf{Necessity of Cluster-Information Enhancement.} Since SDC identifies clusters by the decision graph, it is crucial whether density mountains can be easily distinguished from the decision graph. We refer to the SDC without the cluster-information enhancement process as SDC-NoEnhan. Next, we compare the decision graphs of SDC and SDC-NoEnhan on \emph{overlap1} and \emph{overlap2} datasets with overlapping clusters, as shown in Figure \ref{fig5}. The experimental results show that it is difficult to distinguish different density mountains from the decision graphs of SDC-NoEnhan. However, different density mountains are clearly visible in the decision graphs of SDC. We also calculate the accuracy of SDC and SDC-NoEnhan on both datasets. The accuracy of SDC-NoEnhan is only 0.60 and 0.76, while the accuracy of SDC reaches 0.85 and 0.94. Clearly, the cluster-information enhancement process is essential for SDC.

	\begin{table*}[t]
		\caption{The accuracies of SDC and baseline algorithms on missing datasets}
		\label{table2}
		\label{sample-table}
		\vskip 0.15in
		\setlength{\tabcolsep}{4.7pt}
		\renewcommand{\arraystretch}{0.88}
		\begin{threeparttable}
		\begin{center}
		\begin{small}
		\begin{sc}
		\begin{tabular}{c|cccccccccccccc|c}
			\toprule
			\multicolumn{2}{c}{Dataset} & \emph{Iris} & \emph{Brea} & \emph{Wine} & \emph{Bank} & \emph{Htru} & \emph{Know} & \emph{Bir1} & \emph{Bir2} & \emph{Ove1} & \emph{Ove2} & \emph{Worm} & \emph{Urba} & \emph{Winn} & \textbf{Avg}\\
			\midrule
			\multirow{11}{*}{ARI}
			&\textbf{SDC (Ours)}	&\textbf{0.8}	&\textbf{0.66}	&\textbf{0.52}	&\textbf{0.16}	&\textbf{0.54}	&\textbf{0.26}	&\textbf{0.79}	&0.33	&0.74	&\textbf{0.81}	&\textbf{0.1}	&\textbf{0.42}	&\textbf{0.57}	&\textbf{0.52}	\\
			&MICE	&0.71	&0.49	&0.38	&0.05	&-0.08	&0.15	&0.71	&0.51	&0.78	&0.72	&\textbf{0.1}	&0.33	&0.36	&0.4	\\
			&LRC	&0.52	&0.47	&0.28	&0.04	&-0.08	&0.1	&0.72	&0.49	&0.53	&0.68	&\textbf{0.1}	&0.27	&0.3	&0.34	\\
			&GAIN	&0.7	&0.48	&0.32	&0.04	&-0.04	&0.17	&0.75	&0.49	&0.6	&0.74	&-0.73	&0.33	&0.41	&0.33	\\
			&MDIOT	&0.7	&0.45	&0.28	&0.05	&-0.07	&0.15	&0.75	&0.48	&0.76	&0.73	&\textbf{0.1}	&0.33	&0.41	&0.39	\\
			&VAEAC	&0.75	&0.49	&0.36	&0.05	&-0.08	&0.17	&0.75	&0.59	&0.79	&0.78	&\textbf{0.1}	&0.33	&0.41	&0.42	\\
			&KPOD	&0.65	&0.19	&0.45	&0.06	&0.48	&0.13	&0.46	&0.37	&0.58	&0.32	&0.09	&0.33	&0.41	&0.35	\\
			&TDM	&0.67	&0.45	&0.28	&0.05	&-0.07	&0.14	&0.71	&0.48	&0.78	&0.73	&\textbf{0.1}	&0.32	&0.4	&0.39	\\
			&HyperImpute	&0.67	&0.49	&0.39	&0.05	&-0.08	&0.14	&0.71	&\textbf{0.63}	&0.79	&0.76	&\textbf{0.1}	&0.33	&0.42	&0.42	\\
			&MBFS	&0.7	&0.49	&0.41	&0.05	&-0.08	&0.17	&0.77	&0.6	&\textbf{0.81}	&0.76	&0.09	&0.34	&-\tnote{1}	&0.39	\\
			&DIMV	&0.69	&0.48	&0.34	&0.05	&-0.08	&0.15	&0.73	&0.48	&0.72	&0.74	&\textbf{0.1}	&0.33	&0.39	&0.39	\\
			
			\bottomrule
			
			\toprule
			\multicolumn{2}{c}{Dataset} & \emph{Iris} & \emph{Brea} & \emph{Wine} & \emph{Bank} & \emph{Htru} & \emph{Know} & \emph{Bir1} & \emph{Bir2} & \emph{Ove1} & \emph{Ove2} & \emph{Worm} & \emph{Urba} & \emph{Winn} & \textbf{Avg}\\
			\midrule
			\multirow{11}{*}{NMI}
			&\textbf{SDC (Ours)}	&\textbf{0.79}	&\textbf{0.51}	&\textbf{0.53}	&\textbf{0.25}	&\textbf{0.38}	&\textbf{0.32}	&\textbf{0.91}	&\textbf{0.87}	&0.62	&\textbf{0.86}	&\textbf{0.37}	&\textbf{0.52}	&0.59	&\textbf{0.58}	\\
			&MICE	&0.75	&0.46	&0.44	&0.03	&0.03	&0.19	&0.87	&0.81	&0.68	&0.81	&0.33	&0.5	&0.59	&0.5	\\
			&LRC	&0.53	&0.45	&0.29	&0.03	&0.03	&0.14	&0.87	&0.81	&0.51	&0.8	&0.31	&0.42	&0.47	&0.44	\\
			&GAIN	&0.72	&0.46	&0.38	&0.03	&0.01	&0.22	&0.9	&0.81	&0.55	&0.81	&0.34	&0.51	&\textbf{0.64}	&0.49	\\
			&MDIOT	&0.72	&0.43	&0.36	&0.03	&0.02	&0.19	&0.9	&0.81	&0.67	&0.81	&0.34	&0.51	&0.63	&0.49	\\
			&VAEAC	&0.77	&0.46	&0.42	&0.03	&0.03	&0.22	&0.87	&0.84	&0.69	&0.83	&0.33	&0.5	&0.63	&0.51	\\
			&KPOD	&0.7	&0.23	&0.52	&0.07	&0.36	&0.18	&0.77	&0.76	&0.57	&0.56	&0.33	&0.47	&0.53	&0.47	\\
			&TDM	&0.71	&0.43	&0.36	&0.03	&0.02	&0.17	&0.87	&0.81	&0.69	&0.81	&0.34	&0.5	&0.61	&0.49	\\
			&HyperImpute	&0.72	&0.46	&0.44	&0.03	&0.03	&0.17	&0.87	&0.82	&0.69	&0.82	&0.31	&0.5	&\textbf{0.64}	&0.5	\\
			&MBFS	&0.74	&0.46	&0.44	&0.03	&0.03	&0.22	&0.86	&0.76	&\textbf{0.71}	&0.81	&0.25	&0.51	&-\tnote{1}	&0.45	\\
			&DIMV	&0.73	&0.46	&0.4	&0.03	&0.03	&0.19	&0.88	&0.81	&0.64	&0.81	&0.34	&0.5	&0.55	&0.49	\\
			
			\bottomrule
			
			\toprule
			\multicolumn{2}{c}{Dataset} & \emph{Iris} & \emph{Brea} & \emph{Wine} & \emph{Bank} & \emph{Htru} & \emph{Know} & \emph{Bir1} & \emph{Bir2} & \emph{Ove1} & \emph{Ove2} & \emph{Worm} & \emph{Urba} & \emph{Winn} & \textbf{Avg}\\
			\midrule
			\multirow{11}{*}{PUR}
			&\textbf{SDC (Ours)}	&\textbf{0.94}	&\textbf{0.92}	&\textbf{0.87}	&\textbf{0.83}	&\textbf{0.96}	&\textbf{0.63}	&\textbf{0.9}	&\textbf{0.85}	&0.94	&\textbf{0.92}	&\textbf{0.25}	&0.55	&0.79	&\textbf{0.8}	\\
			&MICE	&0.88	&0.85	&0.7	&0.62	&0.91	&0.53	&0.83	&0.56	&0.94	&0.85	&0.21	&0.69	&0.82	&0.72	\\
			&LRC	&0.74	&0.85	&0.64	&0.6	&0.91	&0.48	&0.83	&0.56	&0.86	&0.82	&0.21	&0.61	&0.72	&0.68	\\
			&GAIN	&0.88	&0.85	&0.67	&0.6	&0.91	&0.54	&0.78	&0.58	&0.88	&0.86	&0.21	&0.7	&0.86	&0.72	\\
			&MDIOT	&0.89	&0.84	&0.66	&0.61	&0.91	&0.54	&0.79	&0.58	&0.94	&0.86	&0.21	&0.69	&0.86	&0.72	\\
			&VAEAC	&0.9	&0.85	&0.69	&0.61	&0.91	&0.56	&0.86	&0.69	&0.94	&0.9	&0.21	&0.68	&0.86	&0.74	\\
			&KPOD	&0.84	&0.73	&0.71	&0.62	&0.94	&0.52	&0.62	&0.46	&0.88	&0.48	&0.22	&0.63	&0.7	&0.64	\\
			&TDM	&0.86	&0.84	&0.66	&0.61	&0.91	&0.52	&0.83	&0.58	&0.94	&0.86	&0.21	&0.68	&0.85	&0.72	\\
			&HyperImpute	&0.85	&0.85	&0.71	&0.61	&0.91	&0.51	&0.83	&0.68	&0.94	&0.88	&0.22	&0.69	&\textbf{0.87}	&0.73	\\
			&MBFS	&0.88	&0.85	&0.74	&0.61	&0.91	&0.53	&0.87	&0.62	&\textbf{0.95}	&0.87	&0.21	&\textbf{0.7}	&-\tnote{1}	&0.67	\\
			&DIMV	&0.87	&0.85	&0.69	&0.62	&0.91	&0.53	&0.84	&0.58	&0.92	&0.86	&0.21	&0.68	&0.74	&0.72	\\
			
			\bottomrule
		\end{tabular}
		\end{sc}
		\end{small}
		\begin{tablenotes}
			\footnotesize
		 	\item[1] Runtime is too long to get results.
		\end{tablenotes}
		\end{center}
		\end{threeparttable}
		\vskip -0.1in
	\end{table*}
	
	\textbf{Necessity of Lightweight Density Calculation.} We refer to the SDC without the lightweight density calculation process as SDC-NoLight. We test the runtime of SDC and SDC-NoLight on two datasets, \emph{birch1} and \emph{birch2}, each containing 100,000 objects. As shown in Figure \ref{fig6}, SDC-NoLight has an extremely long runtime, exceeding 5,000 seconds on both datasets. Conversely, SDC runs in less than 5 seconds. Clearly, the lightweight density calculation process can drastically reduce the time complexity of SDC.
		
	\subsection{Comparison Experiments}
	\textbf{Accuracy Comparison.} The accuracies of SDC and baseline algorithms on 13 missing datasets are recorded in Table \ref{table2}, where the optimal results are bolded. The average accuracy of each algorithm across all datasets is also computed. The experimental results reveal that SDC performs the best. Specifically, for NMI metric, SDC is only inferior to baseline algorithms on \emph{overlap1} and \emph{winnipeg}; For ARI metric, SDC achieves the best results on 11 out of 13 missing datasets; For Purity metric, SDC achieves the best results on 10 out of 13 missing datasets. Although HyperImpute, MBFS, and GAIN have a higher accuracy than SDC on a few datasets, they perform noticeably worse than SDC on the remaining datasets. In terms of average accuracy, SDC outperforms baseline algorithms by at least 13.7\%(NMI), 23.8\%(ARI), and 8.1\%(Purity). Additionally, we compare the performance of SDC and baseline algorithms under different missing rates. Specifically, we construct 7 groups of datasets with missing rates ranging from 10\% to 70\% by removing or completing original values, and then calculate the average accuracy of each algorithm on each group of datasets (Detailed results are provided in Appendix \ref{appendix_missing_acc}). Figure \ref{fig7} (A) shows how these average accuracies (NMI) vary with the missing rate. How these average accuracies (ARI, Purity) vary with the missing rate is provided in Appendix \ref{appendix_average_acc}. The results indicate that, regardless of the evaluation metric, the average accuracies of all algorithms decrease as the missing rate increases. However, SDC's curve consistently remains above the curves of baseline algorithms, indicating that SDC's advantage over baseline algorithms is not affected by the missing rate of datasets.
	
	\textbf{Runtime Comparison.} Figure \ref{fig7} (B) illustrates the runtime of SDC and baseline algorithms on different missing datasets. The results show that VAEAC, MBFS, HyperImpute, and KPOD run very slowly, taking over 500 seconds on large-scale datasets. Especially for MBFS, its runtime is too long to get results on \emph{winnipeg}. TDM, MDIOT, DIMV, MICE, and LRC perform better than the above algorithms, but they still require more than 50 seconds on some datasets. The remaining algorithms (including SDC) run in less than 1 second on most datasets. In summary, the runtime of SDC is entirely acceptable.
	
	\begin{figure*}[t!]
		\vskip 0.1in
		\begin{center}
			\includegraphics[width=0.99\textwidth]{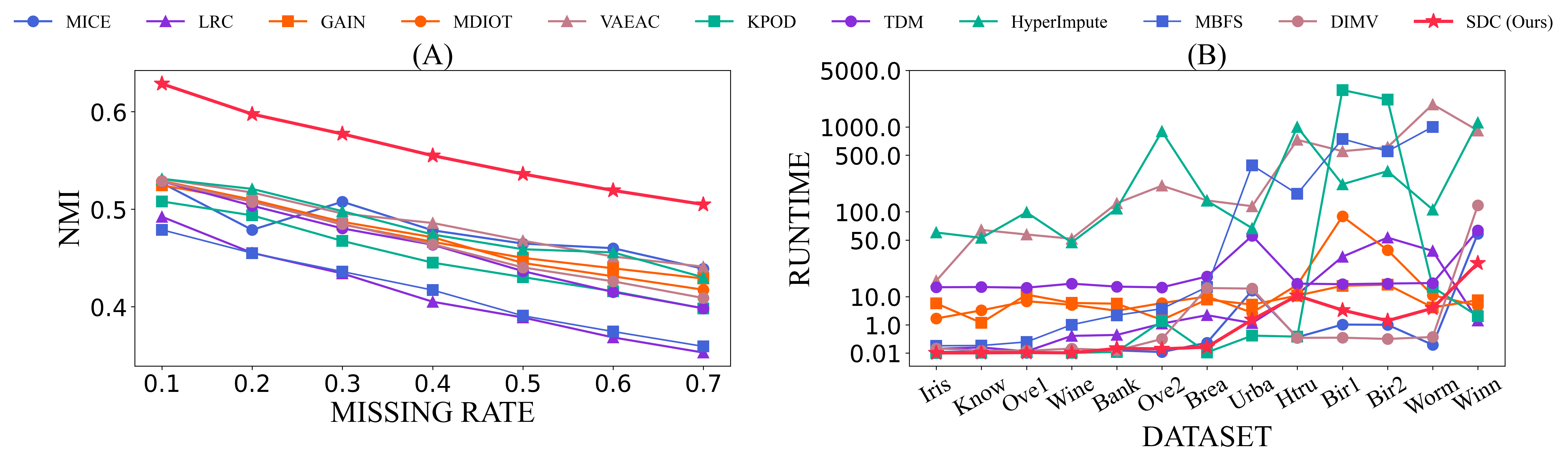}
			\caption{Average accuracy (A) and runtime (B) of SDC and baseline algorithms.}
			\label{fig7}
		\end{center}
		\vskip -0.1in
	\end{figure*}

	\section{Related Work}
	Current algorithms for the clustering task with missing values can be classified into two categories: the first category involves an independent imputation process and clustering process, while the second category involves an alternating loop between the imputation and clustering processes.
	
	\textbf{Category 1.} MICE \cite{JSSv045i03} models features with missing values as functions of other features. GAIN \cite{Yoon2018GAINMD} utilizes a generative adversarial network to impute missing features, where a discriminator tries to distinguish between imputed and observed values. VAEAC \cite{ivanov2018variational} proposes a VAE based neural probabilistic model for distribution conditioning learning and applies it to feature imputation. MDIOT \cite{Muzellec2020MissingDI} assumes that two random subsets in the same dataset share the same distribution and uses optimal transport distance to quantify the distribution differences to impute missing values. MBFS \cite{liu2023improving} captures an optimal subset of features for each object and then fills in missing values based on the found missingness mechanism. TDM \cite{10.5555/3618408.3620181} imputes missing value by transforming two batches of samples into a latent space and matching them distributionally. DIMV \cite{Vu2023ConditionalEW} determines the conditional distribution of a feature that has missing entries, based on the fully observed features. \emph{However, for the above algorithm, the imputation process may not align with clustering principles, potentially resulting in imputed objects that are not conducive to the clustering process, thus increasing the difficulty of clustering. In this work, the proposed SDC directly clusters missing objects through dimension splitting, avoiding the imputation process.}
	
	\textbf{Category 2.} KPOD \cite{Chi2014kPODAM} iteratively performs K-Means and imputation processes. FCMID \cite{ZHANG201651} establishes a Gaussian model for missing values, jointly optimizing the FCM objective function and the maximum likelihood function of missing values. SSC-MC \cite{FAN201736} improves the objective function of SSC to adapt to missing data and iterativly solves out coefficient matrices reflecting cluster membership information. GMMID \cite{10.1145/3408318} executing GMM and imputation processes alternately. CI-clustering \cite{9430134} proposes a local imputation clustering algorithm, first clustering complete objects, then sequentially clustering and imputing incomplete objects. \emph{However, the above algorithms require inputting multiple parameters during the imputation and clustering processes, making it challenging to select appropriate parameter values in real-world analysis scenarios. In this work, the proposed SDC employs decision graphs instead of parameters, allowing users to obtain effective results without needing to specify any parameters.}

\section{Conclusion}
	In this paper, we focus on eliminating input parameters in the clustering task with missing values, and propose an algorithm called SDC. Without any input parameters, SDC can obtain clustering results on missing datasets with the help of decision graphs. In addition, we design the cluster-information enhancement process and the lightweight density calculation process to ensure the effectiveness and low-complexity of SDC. Experimental results show that SDC outperforms baseline algorithms by at least 13.7\%(NMI), 23.8\%(ARI), and 8.1\%(Purity).

\nocite{langley00}
\section*{Impact Statements}
	This paper presents work whose goal is to advance the field of Machine Learning. There are many potential societal consequences of our work, none which we feel must be specifically highlighted here.
\bibliography{reference}
\bibliographystyle{icml2024}

\newpage
\appendix
\onecolumn


\section{The proof of theorems}
\label{appendix_theorems}
\begin{theorem} (i.e. \textbf{Theorem \ref{theorem-1}})
	Let $div(i)$ be $\{ {clu(i)}_{1},{clu(i)}_{2},\cdots,$${clu(i)}_{S(i)}\}$ and $\{{clu}_{1},{clu}_{2},\cdots,{clu}_{S} \}$ be the true clusters in $X$. For $\forall x_{g} \in {clu(i)}_{k}$ and $\forall x_{r} \in {clu(i)}_{l}$, if ${clu(i)}_{k} \neq {clu(i)}_{l}$, then $\nexists{clu}_{p} \in \left\{ {clu}_{1},{clu}_{2},\cdots,{clu}_{S} \right\}$, such that $x_{g},x_{r} \in {clu}_{p}$.
\end{theorem}
\emph{Proof.} 

Let $\delta$ be the minimum inter-cluster distance.

$\because$ For $\forall{clu(i)}_{k},{clu(i)}_{l} \in div(i)$,
${clu(i)}_{k} \neq {clu(i)}_{l}$.

$\therefore$ For $\forall x_{g} \in {clu(i)}_{k}$ and
$\forall x_{r} \in {clu(i)}_{l}$, $\vert x_{g}(i) - x_{r}(i) \vert> \delta$,
where $x_{g}(i)$ and $x_{r}(i)$ are the values of $x_{g}$ and
$x_{r}$ in the $i$-th dimension, respectively.

$\because$
$\left\| x_{g} - x_{r} \right\|_{2} = \sqrt{\left( x_{g}(1) - x_{r}(1) \right)^{2} + \cdots + \left( x_{g}(i) - x_{r}(i) \right)^{2} + \cdots + \left( x_{g}(d) - x_{r}(d) \right)^{2}}$.

$\therefore$ For $\forall x_{g} \in {clu(i)}_{k}$ and
$\forall x_{r} \in {clu(i)}_{l}$,
$\left\| x_{g} - x_{r} \right\|_{2} > \delta$.

$\therefore$ For $\forall{clu(i)}_{k},{clu(i)}_{l} \in div(i)$, the
minimum distance between the objects in ${clu(i)}_{k}$ and the objects
in ${clu(i)}_{l}$ is greater than $\delta$.

$\therefore$ For $\forall{clu(i)}_{k},{clu(i)}_{l} \in div(i)$, the
objects in ${clu(i)}_{k}$ and the objects in ${clu(i)}_{l}$ must not
belong to the same true cluster.

$\therefore$
$\nexists{clu}_{p} \in \left\{ {clu}_{1},{clu}_{2},\cdots,{clu}_{S} \right\}$\emph{,
	such that} $x_{g},x_{r} \in {clu}_{p}$\emph{.}

\rightline{$\blacksquare$}

\begin{theorem}(i.e. \textbf{Theorem \ref{theorem-2}})
	$div(i) \cap div(i - 1) = \big\{ {clu(i)}_{k} \cap {clu(i - 1)}_{l} \big| \forall{clu(i)}_{k} \in div(i),\ \forall{clu(i - 1)}_{l} \in div(i - 1) \big\}$ is the partition intersection between $div(i)$ and $div(i - 1)$. $\left\{{clu}_{1},{clu}_{2},\cdots,{clu}_{S} \right\}$ is the true cluster in $X$. For $\forall x_{g} \in {clu(i)}_{k} \cap {clu(i - 1)}_{l}$, $\forall x_{r} \in {clu(i)}_{p} \cap {clu(i - 1)}_{q}$, if ${clu(i)}_{k} \cap {clu(i - 1)}_{l} \neq {clu(i)}_{p} \cap {clu(i - 1)}_{q}$, then $\nexists{clu}_{f} \in \left\{ {clu}_{1},{clu}_{2},\cdots,{clu}_{S} \right\}$, such that $x_{g},x_{r} \in {clu}_{f}$.
\end{theorem}
\emph{Proof.}

$\because$ $x_{g} \in {clu(i)}_{k} \cap {clu(i - 1)}_{l}$, and
$x_{r} \in {clu(i)}_{p} \cap {clu(i - 1)}_{q}$\emph{.}

$\therefore$ $x_{g} \in {clu(i)}_{k}$, and
$x_{r} \in {clu(i)}_{p}$\emph{.}

$\therefore$ According to Theorem
1, $\nexists{clu}_{f} \in \left\{ {clu}_{1},{clu}_{2},\cdots,{clu}_{S} \right\}$\emph{,
	such that} $x_{g},x_{r} \in {clu}_{f}$\emph{.}

\rightline{$\blacksquare$}

\begin{theorem}(i.e. \textbf{Theorem \ref{theorem-4}})
	For $\forall x_{i},x_{j} \in X$, if $x_{j} \in \Gamma\left(x_{i}\right)$, then $x_{j} \in \Psi\left( x_{i}\right)$.
\end{theorem}	
\emph{Proof.} 

Let $\theta$ be the virtual object. According to trigonometric inequality, for objects $\theta$, $x_{i}$, and $x_{j}$, the inequality $\| x_{i} - \theta\|_{2} - \| x_{i} - x_{j}\|_{2} \leq \| x_{j} - \theta\|_{2} \leq \| x_{i} - \theta\|_{2} + \| x_{i} - x_{j}\|_{2}$ holds.

$\because$ $x_{j} \in \Gamma\left( x_{i} \right)$.

$\therefore$ According to Definition \ref{def-3},
$\| x_{i} - x_{j}\|_{2} \leq R$.

$\therefore$
$\| x_{i} - \theta\|_{2} - R \leq \| x_{j} - \theta\|_{2} \leq \| x_{i} - \theta\|_{2} + R$.

$\therefore$ According to Definition \ref{def-4},
$x_{j} \in \Psi\left( x_{i} \right)$.

\rightline{$\blacksquare$}

\begin{theorem}(i.e. \textbf{Theorem \ref{theorem-5}})
	For $\forall x_{j} \in X$, if $x_{j} \in \Psi_{i}(X)$, then $\Gamma\left( x_{j} \right) \subset \Psi_{i - 1}(X) \cup \Psi_{i}(X) \cup \Psi_{i + 1}(X)$.
\end{theorem}	
\emph{Proof.} 

To prove that $\Gamma\left( x_{j} \right) \subset \Psi_{i - 1}(X) \cup \Psi_{i}(X) \cup \Psi_{i + 1}(X)$,	simply prove that ``if $\forall x_{g} \in \Gamma\left( x_{j} \right)$, then $x_{g} \in \Psi_{i - 1}(X) \cup \Psi_{i}(X) \cup \Psi_{i + 1}(X)$''. Let $\theta$ be the virtual object. According to trigonometric inequality, for objects $\theta$, $x_{j}$, and $x_{g}$, the inequality $\| x_{j} - \theta\|_{2} - \| x_{j} - x_{g}\|_{2} \leq \| x_{g} - \theta\|_{2} \leq \| x_{j} - \theta\|_{2} + \| x_{j} - x_{g}\|_{2}$ holds.

$\because$ $x_{j} \in \Psi_{i}(X)$.

$\therefore$ According to Definition \ref{def-5},
$(i - 1)R < \| x_{j} - \theta\|_{2} \leq iR$.

$\because$ $x_{g} \in \Gamma\left( x_{j} \right)$.

$\therefore$ According to Definition \ref{def-3},
$\| x_{g} - x_{j}\|_{2} \leq R$.

$\therefore$
$(i - 2)R < \| x_{j} - \theta\|_{2} - \| x_{g} - x_{j}\|_{2}$, and
$\| x_{j} - \theta\|_{2} + \| x_{g} - x_{j}\|_{2} \leq (i + 1)R$.

$\therefore$ $(i - 2)R < \| x_{g} - \theta\|_{2} \leq (i + 1)R$.

$\therefore$ According to Definition \ref{def-5},
$x_{g} \in \Psi_{i - 1}(X) \cup \Psi_{i}(X) \cup \Psi_{i + 1}(X)$.

$\therefore$
$\Gamma\left( x_{j} \right) \subseteq \Psi_{i - 1}(X) \cup \Psi_{i}(X) \cup \Psi_{i + 1}(X)$.

\rightline{$\blacksquare$}

\begin{theorem}(i.e. \textbf{Theorem \ref{theorem-6}})
	For $\forall x_{i},x_{j} \in X$, if $x_{j} \in \Gamma\left( x_{i} \right)$, then $\left| D_{i} - D_{j} \right| \leq \sqrt{d}R$.
\end{theorem}	
\emph{Proof.}

$\because$ $x_{j} \in \Gamma\left( x_{i} \right)$.

$\therefore$ According to Definition \ref{def-3},
$\| x_{j} - x_{i}\|_{2} \leq R$.

$\therefore$
$\sqrt{\left( x_{j}(1) - x_{i}(1) \right)^{2} + \left( x_{j}(2) - x_{i}(2) \right)^{2} + \cdots + \left( x_{j}(d) - x_{i}(d) \right)^{2}} \leq R$, where $x_{j}(1)$ and $x_{i}(1)$ are the values of $x_{j}$ and
$x_{i}$ in the first dimension, and so on.

$\therefore$
$\left( x_{j}(1) - x_{i}(1) \right)^{2} + \left( x_{j}(2) - x_{i}(2) \right)^{2} + \cdots + \left( x_{j}(d) - x_{i}(d) \right)^{2} \leq R^{2} = \frac{dR^{2}}{d}$

$\therefore$
$\left( \left( x_{j}(1) - x_{i}(1) \right)^{2} - \frac{R^{2}}{d} \right) + \left( \left( x_{j}(2) - x_{i}(2) \right)^{2} - \frac{R^{2}}{d} \right) + \cdots + \left( \left( x_{j}(d) - x_{i}(d) \right)^{2} - \frac{R^{2}}{d} \right) \leq 0$.

$\therefore$
$\left( \left( x_{j}(1) - x_{i}(1) \right) + \frac{R}{\sqrt{d}} \right) \left( \left( x_{j}(1) - x_{i}(1) \right) - \frac{R}{\sqrt{d}} \right) + \left( \left( x_{j}(2) - x_{i}(2) \right) + \frac{R}{\sqrt{d}} \right) \left( \left( x_{j}(2) - x_{i}(2) \right) - \frac{R}{\sqrt{d}} \right) + \cdots + \left( \left( x_{j}(d) - x_{i}(d) \right) + \frac{R}{\sqrt{d}} \right)\left( \left( x_{j}(d) - x_{i}(d) \right) - \frac{R}{\sqrt{d}} \right) \leq 0$.

$\because$ The theorem only considers relative distances between
objects, so pairs of objects with the same relative distance are
equivalent to each other.

$\therefore$ It can be assumed that for $\forall s \leq d$,
$\left( x_{j}(s) - x_{i}(s) \right) \geq 0$.

$\therefore$ For $\forall s \leq d$,
$\left( \left( x_{j}(s) - x_{i}(s) \right) + \frac{R}{\sqrt{d}} \right) > 0$.

$\therefore$
$\left( \left( x_{j}(1) - x_{i}(1) \right) - \frac{R}{\sqrt{d}} \right) + \left( \left( x_{j}(2) - x_{i}(2) \right) - \frac{R}{\sqrt{d}} \right) + \cdots + \left( \left( x_{j}(d) - x_{i}(d) \right) - \frac{R}{\sqrt{d}} \right) \leq 0$.

$\therefore$
$\left( x_{j}(1) - x_{i}(1) \right) + \left( x_{j}(2) - x_{i}(2) \right) + \cdots + \left( x_{j}(d) - x_{i}(d) \right) \leq \sqrt{d}R$.

$\therefore$
$\left( x_{j}(1) + x_{j}(2) + \cdots + x_{j}(d) \right) - \left( x_{i}(1) + x_{i}(2) + \cdots + x_{i}(d) \right) \leq \sqrt{d}R$.

$\therefore$ $|D_{j} - D_{i}| \leq \sqrt{d}R$\emph{.}

\rightline{$\blacksquare$}

\section{Time complexity analysis of the lightweight density calculation process}
\label{appendix_time_density}

The time complexity analysis of the lightweight density calculation
process, which first divides public-annular regions and then calculates the
density, is discussed below:

\begin{itemize}
	\item
	\textbf{Step1 (dividing public-annular regions):} The step first calculates
	the distance from all objects to the virtual object and sorts them by
	distance, and then traverses the objects only once to divide them into
	$T$ public-annular regions, so the time complexity is
	$O\left( N\log N + N + N \right)$.
	\item
	\textbf{Step2 (calculating the density):} The average number
	of objects in each public-annular region is $N/T$. When ``batch''
	calculating the density of objects in $\Psi_{i}(X)$, the step first
	calculates the sums of dimensional values of the objects in
	$\Psi_{i - 1}(X) \cup \Psi_{i}(X) \cup \Psi_{i + 1}(X)$, and sorts
	the objects by the sums, with a time complexity of
	$O\left( \frac{3N}{T}\log\left( \frac{3N}{T} \right) \right)$. Next, the step
	looks for objects in
	$\Psi_{i - 1}(X) \cup \Psi_{i}(X) \cup \Psi_{i + 1}(X)$ that meet
	the dimensionality relationship condition with the objects in
	$\Psi_{i}(X)$, and calculates the distance between these objects to
	obtain the density, with a time complexity of
	$O\left( \frac{NS}{T} \right)$,
	where $S$ is the average number of objects that meet the
	dimensionality relationship condition with each object in
	$\Psi_{i}(X)$, $S \ll N$. Therefore, the time complexity of
	calculating the density of all objects in $T$ public-annular regions is
	$O\left( 3N\log\left( \frac{3N}{T} \right) + NS \right)$, where
	$\frac{3N}{T},S \ll N$.
\end{itemize}

In summary, the time complexity of the lightweight density calculation process is $O( N\log N + 3N\log\left( \frac{3N}{T} \right) + (S + 2)N )$. Since $\frac{3N}{T},S \ll N$, this complexity is approximate to $O\left( N\log N \right)$.

\section{Time complexity analysis of SDC}
\label{appendix_time_SDC}

The lightweight density calculation process is invoked by the clustering process and the cluster-information enhancement process, so we discuss the time complexity of SDC in terms of the clustering process and the cluster-information enhancement process.

\begin{itemize}
	\item
	\textbf{The clustering process:} (1) SDC calculates the density and draws the density decision graphs for $d$ single-dimensional datasets with a time complexity of $O\left( dN\log N + dN \right)$. (2) After the user determines the partition thresholds (\emph{i.e.}, the boundaries of density mountains) from the decision graphs, according to these thresholds, SDC traverses the objects once to divide each single-dimensional dataset into different clusters, with a time complexity of $O(dN)$. (3) When performing the partition intersection, SDC uses the dictionary data structure to store the labels of each object in different cluster-partitions to find the fusion clusters, so the time complexity is $O(dN)$. (4) When merging missing objects, SDC only needs to additionally traverse the objects in each single-dimensional dataset once, so the time complexity is $O(dN)$. Ultimately, the time complexity of the clustering process is $O\left( dN\log N + 4dN \right)$.
	\item
	\textbf{The cluster-information enhancement process:} (1) SDC first calculates the density of the objects in $X$ in order to recognize the low-density objects, with a time complexity of $O\left( N\log N \right)$. (2) Since the number of low-density objects is not significantly less than the number of objects $N$, the number of low-density objects belongs to the same order of magnitude as $N$. Therefore, SDC finds 5 nearest neighbors for each low-density object with a time complexity of $O\left( N\log N \right)$. (3) SDC calculates the gravitational force between each low-density object and its 5 nearest neighbors with a time complexity of $O(5N)$. (4) SDC moves each low-density object with a time complexity of $O(N)$. Ultimately, the time complexity of the cluster-information enhancement process is $O\left( 2N\log N + 6N \right)$.
\end{itemize}

In summary, the time complexity of SDC is $O( (d + 2)N\log N + 4dN + 6N )$, that is,	$O( (d + 2)N\log N )$.

\section{The parameter intervals of baseline algorithms}
\label{appendix_parameter}
In this paper, the parameter intervals for the baseline algorithms are set to their recommended intervals in their original papers, as follows: for TDM, both $T$ and $K$ range from 1 to 4; for DIMV, $ \alpha \in [0, 0.01, 0.1, 1, 10, 100]$; for MBFS, \emph{ntree} is the parameter of MissForest and ranges from 10 to 100; for MICE, $strategy \in $ {[}\textquotesingle mean\textquotesingle, \textquotesingle median\textquotesingle, \textquotesingle most\_frequent\textquotesingle, \textquotesingle constant\textquotesingle{]}, $order \in $ {[}\textquotesingle ascending\textquotesingle, \textquotesingle descending\textquotesingle, \textquotesingle roman\textquotesingle, \textquotesingle arabic\textquotesingle{]}, $n\_nearest\_features \in [1, d-1]$; for LRC, \(p \in \lbrack 0.1,\ 2.0\rbrack\), \(r \in \lbrack 0.5,1.5,2.5,3.5,4.5\rbrack\); for KPOD, \(tol \in \lbrack 0.01,0.2\rbrack\); for GAIN, \(h \in \lbrack 0.1,1\rbrack\), \(\alpha \in \lbrack 10,100\rbrack\), \(k\in[1, 30]\); for VAEAC, \(num \in \lbrack 1,10\rbrack\); for MDIOT, \(scale \in \lbrack 0.1,0.9\rbrack\), \(\varepsilon \in \lbrack 0.1,1\rbrack\), $quantile \in [0.1, 0.9]$.

\newpage
\section{How the average accuracies (ARI, Purity) vary with the missing rate}
\label{appendix_average_acc}
\begin{figure}[hb]
	\centering
	\includegraphics[width=0.95\columnwidth]{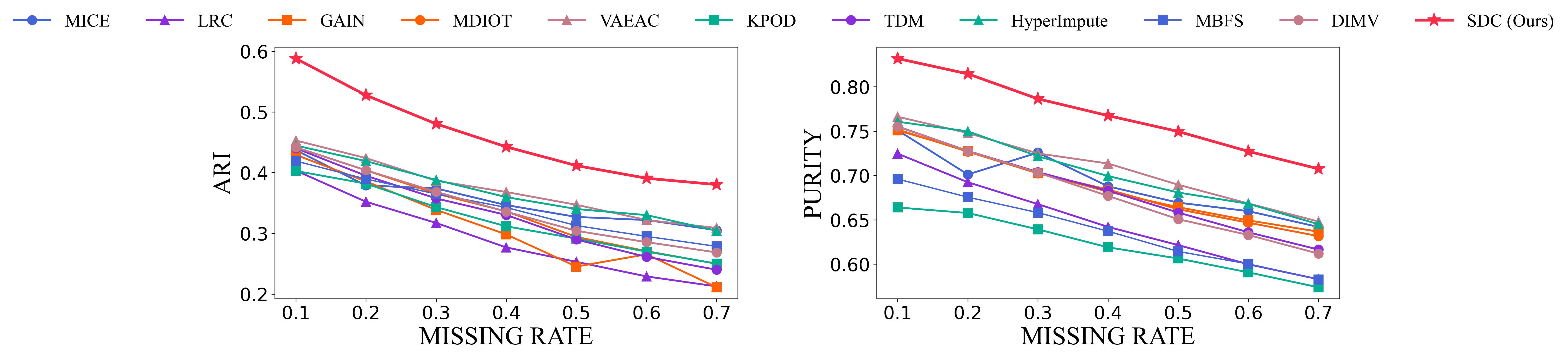}
\end{figure}

\section{The accuracies of all algorithms on 13 datasets with missing rates ranging from 10\% to 70\%}
\label{appendix_missing_acc}

\begin{table*}[ht!]
	\caption{The accuracies of SDC and baseline algorithms on 13 missing datasets with a missing rate of 0.1}
	\label{table3}
	\vskip 0.15in
	\setlength{\tabcolsep}{4.7pt}
	\begin{threeparttable}
	\begin{center}
		\begin{small}
			\begin{sc}
				\begin{tabular}{c|cccccccccccccc|c}
					\toprule
					\multicolumn{2}{c}{Dataset} & \emph{Iris} & \emph{Brea} & \emph{Wine} & \emph{Bank} & \emph{Htru} & \emph{Know} & \emph{Bir1} & \emph{Bir2} & \emph{Ove1} & \emph{Ove2} & \emph{Worm} & \emph{Urba} & \emph{Winn} & \textbf{Avg}\\
					\midrule
					\multirow{11}{*}{ARI}
					&\textbf{SDC (Ours)}	&\textbf{0.86}	&\textbf{0.66}	&\textbf{0.58}	&\textbf{0.18}	&0.53	&\textbf{0.31}	&\textbf{0.79}	&0.8	&\textbf{0.86}	&\textbf{0.81}	&\textbf{0.25}	&\textbf{0.42}	&\textbf{0.59}	&\textbf{0.59}	\\
					&MICE	&0.72	&0.49	&0.36	&0.05	&-0.08	&0.17	&0.71	&0.75	&0.85	&0.72	&\textbf{0.25}	&0.33	&0.36	&0.44	\\
					&LRC	&0.58	&0.49	&0.35	&0.05	&-0.08	&0.14	&0.72	&0.74	&0.74	&0.68	&0.24	&0.27	&0.33	&0.4	\\
					&GAIN	&0.69	&0.49	&0.36	&0.05	&-0.07	&0.16	&0.75	&0.75	&0.81	&0.74	&0.11	&0.33	&0.41	&0.43	\\
					&MDIOT	&0.7	&0.49	&0.35	&0.05	&-0.08	&0.17	&0.75	&0.75	&0.85	&0.73	&0.24	&0.33	&0.41	&0.44	\\
					&VAEAC	&0.71	&0.49	&0.35	&0.05	&-0.08	&0.15	&0.75	&\textbf{0.86}	&0.85	&0.78	&\textbf{0.25}	&0.33	&0.41	&0.45	\\
					&KPOD	&0.59	&0.2	&0.54	&0.06	&\textbf{0.54}	&0.17	&0.46	&0.59	&0.82	&0.32	&0.21	&0.33	&0.41	&0.4	\\
					&TDM	&0.71	&0.49	&0.36	&0.05	&-0.08	&0.18	&0.71	&0.75	&0.85	&0.73	&0.24	&0.32	&0.41	&0.44	\\
					&HyperImpute	&0.73	&0.49	&0.37	&0.05	&-0.08	&0.17	&0.71	&\textbf{0.86}	&0.85	&0.76	&0.24	&0.33	&0.4	&0.45	\\
					&MBFS	&0.73	&0.49	&0.35	&0.05	&-0.08	&0.15	&0.77	&0.8	&0.85	&0.76	&\textbf{0.25}	&0.34	&-\tnote{1}	&0.42	\\
					&DIMV	&0.72	&0.49	&0.36	&0.05	&-0.08	&0.16	&0.73	&0.75	&0.85	&0.74	&\textbf{0.25}	&0.33	&0.41	&0.44	\\
					
					\bottomrule
					
					\toprule
					\multicolumn{2}{c}{Dataset} & \emph{Iris} & \emph{Brea} & \emph{Wine} & \emph{Bank} & \emph{Htru} & \emph{Know} & \emph{Bir1} & \emph{Bir2} & \emph{Ove1} & \emph{Ove2} & \emph{Worm} & \emph{Urba} & \emph{Winn} & \textbf{Avg}\\
					\midrule
					\multirow{11}{*}{NMI}
					&\textbf{SDC (Ours)}	&\textbf{0.84}	&\textbf{0.52}	&\textbf{0.61}	&\textbf{0.28}	&0.4	&\textbf{0.38}	&\textbf{0.91}	&\textbf{0.95}	&0.76	&\textbf{0.86}	&\textbf{0.53}	&\textbf{0.52}	&0.61	&\textbf{0.63}	\\
					&MICE	&0.75	&0.46	&0.42	&0.03	&0.03	&0.22	&0.87	&0.9	&0.76	&0.81	&0.5	&0.5	&0.59	&0.53	\\
					&LRC	&0.59	&0.46	&0.39	&0.03	&0.03	&0.2	&0.87	&0.9	&0.69	&0.8	&0.49	&0.42	&0.52	&0.49	\\
					&GAIN	&0.71	&0.46	&0.41	&0.03	&0.02	&0.22	&0.9	&0.9	&0.72	&0.81	&0.5	&0.51	&\textbf{0.64}	&0.53	\\
					&MDIOT	&0.71	&0.46	&0.41	&0.03	&0.03	&0.23	&0.9	&0.9	&0.76	&0.81	&0.5	&0.51	&0.63	&0.53	\\
					&VAEAC	&0.74	&0.46	&0.41	&0.03	&0.03	&0.2	&0.87	&0.94	&0.76	&0.83	&0.5	&0.5	&0.63	&0.53	\\
					&KPOD	&0.63	&0.23	&0.6	&0.06	&\textbf{0.41}	&0.22	&0.77	&0.85	&\textbf{0.78}	&0.56	&0.49	&0.47	&0.55	&0.51	\\
					&TDM	&0.74	&0.46	&0.41	&0.03	&0.03	&0.24	&0.87	&0.9	&0.76	&0.81	&0.5	&0.5	&0.63	&0.53	\\
					&HyperImpute	&0.76	&0.46	&0.43	&0.03	&0.03	&0.23	&0.87	&0.94	&0.76	&0.82	&0.5	&0.5	&0.62	&0.53	\\
					&MBFS	&0.76	&0.46	&0.41	&0.03	&0.03	&0.22	&0.86	&0.9	&0.76	&0.81	&0.49	&0.51	&-\tnote{1}	&0.48	\\
					&DIMV	&0.75	&0.46	&0.42	&0.03	&0.03	&0.21	&0.88	&0.9	&0.76	&0.81	&0.5	&0.5	&0.63	&0.53	\\
					
					\bottomrule
					
					\toprule
					\multicolumn{2}{c}{Dataset} & \emph{Iris} & \emph{Brea} & \emph{Wine} & \emph{Bank} & \emph{Htru} & \emph{Know} & \emph{Bir1} & \emph{Bir2} & \emph{Ove1} & \emph{Ove2} & \emph{Worm} & \emph{Urba} & \emph{Winn} & \textbf{Avg}\\
					\midrule
					\multirow{11}{*}{PUR}
					&\textbf{SDC (Ours)}	&\textbf{0.95}	&\textbf{0.92}	&\textbf{0.9}	&\textbf{0.85}	&\textbf{0.96}	&\textbf{0.66}	&\textbf{0.9}	&\textbf{0.95}	&\textbf{0.97}	&\textbf{0.92}	&\textbf{0.47}	&0.55	&0.82	&\textbf{0.83}	\\
					&MICE	&0.89	&0.85	&0.7	&0.61	&0.91	&0.54	&0.83	&0.76	&0.96	&0.85	&0.35	&0.69	&0.82	&0.75	\\
					&LRC	&0.79	&0.85	&0.69	&0.61	&0.91	&0.51	&0.83	&0.76	&0.93	&0.82	&0.34	&0.61	&0.76	&0.72	\\
					&GAIN	&0.88	&0.85	&0.7	&0.61	&0.91	&0.53	&0.78	&0.78	&0.95	&0.86	&0.34	&\textbf{0.7}	&\textbf{0.87}	&0.75	\\
					&MDIOT	&0.88	&0.85	&0.69	&0.61	&0.91	&0.55	&0.79	&0.79	&0.96	&0.86	&0.34	&0.69	&0.86	&0.75	\\
					&VAEAC	&0.88	&0.85	&0.69	&0.61	&0.91	&0.53	&0.86	&0.88	&0.96	&0.9	&0.34	&0.68	&0.86	&0.77	\\
					&KPOD	&0.79	&0.73	&0.77	&0.62	&0.95	&0.54	&0.62	&0.55	&0.95	&0.48	&0.31	&0.63	&0.71	&0.67	\\
					&TDM	&0.88	&0.85	&0.69	&0.61	&0.91	&0.55	&0.83	&0.78	&0.96	&0.86	&0.34	&0.68	&0.86	&0.75	\\
					&HyperImpute	&0.89	&0.85	&0.7	&0.61	&0.91	&0.55	&0.83	&0.89	&0.96	&0.88	&0.34	&0.69	&0.85	&0.77	\\
					&MBFS	&0.89	&0.85	&0.69	&0.61	&0.91	&0.53	&0.87	&0.81	&0.96	&0.87	&0.35	&\textbf{0.7}	&-\tnote{1}	&0.7	\\
					&DIMV	&0.89	&0.85	&0.7	&0.61	&0.91	&0.53	&0.84	&0.78	&0.96	&0.86	&0.35	&0.68	&0.85	&0.75	\\
					
					\bottomrule
				\end{tabular}
			\end{sc}
		\end{small}
		\begin{tablenotes}
			\footnotesize
			\item[1] Runtime is too long to get results.
		\end{tablenotes}
	\end{center}
	\end{threeparttable}
	\vskip -0.1in
\end{table*}

\begin{table*}
	\caption{The accuracies of SDC and baseline algorithms on 13 missing datasets with a missing rate of 0.2}
	\label{table}
	\vskip 0.15in
	\setlength{\tabcolsep}{4.7pt}
	\begin{threeparttable}
	\begin{center}
		\begin{small}
			\begin{sc}
				\begin{tabular}{c|cccccccccccccc|c}
					\toprule
					\multicolumn{2}{c}{Dataset} & \emph{Iris} & \emph{Brea} & \emph{Wine} & \emph{Bank} & \emph{Htru} & \emph{Know} & \emph{Bir1} & \emph{Bir2} & \emph{Ove1} & \emph{Ove2} & \emph{Worm} & \emph{Urba} & \emph{Winn} & \textbf{Avg}\\
					\midrule
					\multirow{11}{*}{ARI}
					&\textbf{SDC (Ours)}	&\textbf{0.8}	&\textbf{0.66}	&\textbf{0.56}	&\textbf{0.18}	&0.53	&\textbf{0.3}	&0.6	&0.54	&0.81	&\textbf{0.69}	&0.19	&\textbf{0.42}	&\textbf{0.58}	&\textbf{0.53}	\\
					&MICE	&0.71	&0.49	&0.37	&0.05	&-0.08	&0.17	&0.54	&0.62	&0.82	&0.62	&\textbf{0.21}	&0.04	&0.36	&0.38	\\
					&LRC	&0.52	&0.47	&0.32	&0.04	&-0.08	&0.12	&0.57	&0.6	&0.64	&0.59	&\textbf{0.21}	&0.26	&0.33	&0.35	\\
					&GAIN	&0.7	&0.48	&0.35	&0.05	&-0.06	&0.19	&0.6	&0.6	&0.78	&0.62	&-0.06	&0.33	&0.41	&0.38	\\
					&MDIOT	&0.7	&0.45	&0.35	&0.05	&-0.08	&0.19	&0.6	&0.6	&0.82	&0.62	&\textbf{0.21}	&0.33	&0.41	&0.4	\\
					&VAEAC	&0.75	&0.49	&0.35	&0.05	&-0.08	&0.22	&0.59	&0.71	&0.82	&0.66	&\textbf{0.21}	&0.32	&0.41	&0.42	\\
					&KPOD	&0.65	&0.19	&0.53	&0.06	&\textbf{0.55}	&0.17	&0.39	&0.5	&0.73	&0.28	&0.18	&0.32	&0.41	&0.38	\\
					&TDM	&0.67	&0.45	&0.34	&0.05	&-0.08	&0.17	&0.54	&0.6	&0.82	&0.62	&\textbf{0.21}	&0.32	&0.41	&0.39	\\
					&HyperImpute	&0.67	&0.49	&0.39	&0.05	&-0.08	&0.2	&0.54	&\textbf{0.76}	&0.82	&0.66	&\textbf{0.21}	&0.33	&0.41	&0.42	\\
					&MBFS	&0.7	&0.49	&0.39	&0.05	&-0.08	&0.15	&\textbf{0.62}	&0.68	&\textbf{0.83}	&\textbf{0.69}	&\textbf{0.21}	&0.33	&-\tnote{1}	&0.39	\\
					&DIMV	&0.69	&0.48	&0.37	&0.05	&-0.08	&0.17	&0.56	&0.61	&0.8	&0.62	&\textbf{0.21}	&0.34	&0.44	&0.4	\\
					
					\bottomrule
					
					\toprule
					\multicolumn{2}{c}{Dataset} & \emph{Iris} & \emph{Brea} & \emph{Wine} & \emph{Bank} & \emph{Htru} & \emph{Know} & \emph{Bir1} & \emph{Bir2} & \emph{Ove1} & \emph{Ove2} & \emph{Worm} & \emph{Urba} & \emph{Winn} & \textbf{Avg}\\
					\midrule
					\multirow{11}{*}{NMI}
					&\textbf{SDC (Ours)}	&\textbf{0.79}	&\textbf{0.51}	&\textbf{0.59}	&\textbf{0.27}	&0.38	&\textbf{0.37}	&\textbf{0.85}	&\textbf{0.91}	&0.68	&\textbf{0.8}	&\textbf{0.49}	&\textbf{0.52}	&0.61	&\textbf{0.6}	\\
					&MICE	&0.75	&0.46	&0.43	&0.03	&0.03	&0.22	&0.81	&0.85	&0.73	&0.76	&0.46	&0.11	&0.59	&0.48	\\
					&LRC	&0.53	&0.45	&0.35	&0.03	&0.03	&0.18	&0.81	&0.85	&0.59	&0.75	&0.45	&0.4	&0.5	&0.46	\\
					&GAIN	&0.72	&0.46	&0.4	&0.03	&0.02	&0.24	&\textbf{0.85}	&0.85	&0.69	&0.75	&0.46	&0.51	&\textbf{0.64}	&0.51	\\
					&MDIOT	&0.72	&0.43	&0.41	&0.03	&0.02	&0.24	&\textbf{0.85}	&0.85	&0.73	&0.76	&0.46	&0.5	&0.63	&0.51	\\
					&VAEAC	&0.77	&0.46	&0.41	&0.03	&0.03	&0.27	&0.81	&0.85	&0.73	&0.77	&0.46	&0.5	&0.63	&0.52	\\
					&KPOD	&0.7	&0.23	&\textbf{0.59}	&0.07	&\textbf{0.41}	&0.22	&0.73	&0.81	&0.7	&0.52	&0.45	&0.46	&0.54	&0.49	\\
					&TDM	&0.71	&0.43	&0.4	&0.03	&0.03	&0.23	&0.82	&0.85	&0.73	&0.76	&0.46	&0.49	&0.62	&0.5	\\
					&HyperImpute	&0.72	&0.46	&0.44	&0.03	&0.03	&0.26	&0.82	&\textbf{0.91}	&0.73	&0.78	&0.46	&0.51	&0.63	&0.52	\\
					&MBFS	&0.74	&0.46	&0.44	&0.03	&0.03	&0.2	&0.77	&0.82	&\textbf{0.74}	&0.74	&0.45	&0.5	&-\tnote{1}	&0.46	\\
					&DIMV	&0.73	&0.46	&0.42	&0.03	&0.03	&0.22	&0.82	&0.85	&0.71	&0.76	&0.46	&0.5	&0.62	&0.51	\\
					
					\bottomrule
					
					\toprule
					\multicolumn{2}{c}{Dataset} & \emph{Iris} & \emph{Brea} & \emph{Wine} & \emph{Bank} & \emph{Htru} & \emph{Know} & \emph{Bir1} & \emph{Bir2} & \emph{Ove1} & \emph{Ove2} & \emph{Worm} & \emph{Urba} & \emph{Winn} & \textbf{Avg}\\
					\midrule
					\multirow{11}{*}{PUR}
					&\textbf{SDC (Ours)}	&\textbf{0.94}	&\textbf{0.92}	&\textbf{0.91}	&\textbf{0.84}	&\textbf{0.96}	&\textbf{0.66}	&\textbf{0.81}	&\textbf{0.9}	&0.95	&\textbf{0.87}	&\textbf{0.48}	&0.54	&0.81	&\textbf{0.81}	\\
					&MICE	&0.88	&0.85	&0.69	&0.61	&0.91	&0.55	&0.73	&0.65	&0.95	&0.81	&0.32	&0.34	&0.82	&0.7	\\
					&LRC	&0.74	&0.85	&0.68	&0.61	&0.91	&0.5	&0.74	&0.65	&0.9	&0.78	&0.32	&0.59	&0.75	&0.69	\\
					&GAIN	&0.88	&0.85	&0.69	&0.61	&0.91	&0.57	&0.67	&0.67	&0.94	&0.8	&0.32	&\textbf{0.7}	&\textbf{0.87}	&0.73	\\
					&MDIOT	&0.89	&0.84	&0.69	&0.61	&0.91	&0.56	&0.67	&0.67	&0.95	&0.81	&0.31	&0.69	&0.86	&0.73	\\
					&VAEAC	&0.9	&0.85	&0.69	&0.61	&0.91	&0.58	&0.77	&0.74	&0.95	&0.84	&0.32	&0.68	&\textbf{0.87}	&0.75	\\
					&KPOD	&0.84	&0.73	&0.76	&0.62	&0.95	&0.54	&0.58	&0.51	&0.93	&0.46	&0.31	&0.62	&0.7	&0.66	\\
					&TDM	&0.86	&0.84	&0.69	&0.61	&0.91	&0.55	&0.73	&0.67	&0.95	&0.81	&0.31	&0.67	&0.85	&0.73	\\
					&HyperImpute	&0.85	&0.85	&0.71	&0.61	&0.91	&0.56	&0.73	&0.83	&0.95	&0.85	&0.33	&0.69	&\textbf{0.87}	&0.75	\\
					&MBFS	&0.88	&0.85	&0.72	&0.61	&0.91	&0.52	&0.78	&0.7	&\textbf{0.96}	&0.84	&0.32	&0.69	&-\tnote{1}	&0.68	\\
					&DIMV	&0.87	&0.85	&0.7	&0.61	&0.91	&0.54	&0.74	&0.66	&0.95	&0.81	&0.32	&0.67	&0.82	&0.73	\\
					
					\bottomrule
				\end{tabular}
			\end{sc}
		\end{small}
		\begin{tablenotes}
			\footnotesize
			\item[1] Runtime is too long to get results.
		\end{tablenotes}
	\end{center}
	\end{threeparttable}
	\vskip -0.1in
\end{table*}

\begin{table*}
	\caption{The accuracies of SDC and baseline algorithms on 13 missing datasets with a missing rate of 0.3}
	\label{table4}
	\vskip 0.15in
	\setlength{\tabcolsep}{4.7pt}
	\begin{threeparttable}
	\begin{center}
		\begin{small}
			\begin{sc}
				\begin{tabular}{c|cccccccccccccc|c}
					\toprule
					\multicolumn{2}{c}{Dataset} & \emph{Iris} & \emph{Brea} & \emph{Wine} & \emph{Bank} & \emph{Htru} & \emph{Know} & \emph{Bir1} & \emph{Bir2} & \emph{Ove1} & \emph{Ove2} & \emph{Worm} & \emph{Urba} & \emph{Winn} & \textbf{Avg}\\
					\midrule
					\multirow{11}{*}{ARI}
					&\textbf{SDC (Ours)}	&\textbf{0.79}	&\textbf{0.64}	&\textbf{0.56}	&\textbf{0.16}	&\textbf{0.54}	&\textbf{0.29}	&0.44	&0.33	&0.78	&0.56	&0.17	&\textbf{0.41}	&\textbf{0.58}	&\textbf{0.48}	\\
					&MICE	&0.71	&0.49	&0.37	&0.05	&-0.08	&0.15	&0.41	&0.51	&0.8	&0.53	&\textbf{0.18}	&0.33	&0.36	&0.37	\\
					&LRC	&0.48	&0.47	&0.31	&0.04	&-0.08	&0.11	&0.43	&0.49	&0.65	&0.48	&0.17	&0.25	&0.31	&0.32	\\
					&GAIN	&0.67	&0.49	&0.33	&0.04	&-0.05	&0.16	&\textbf{0.49}	&0.49	&0.7	&0.53	&-0.2	&0.33	&0.42	&0.34	\\
					&MDIOT	&0.67	&0.45	&0.32	&0.05	&-0.08	&0.16	&0.48	&0.48	&0.79	&0.53	&0.17	&0.33	&0.41	&0.37	\\
					&VAEAC	&0.74	&0.49	&0.35	&0.05	&-0.08	&0.14	&0.45	&0.59	&0.8	&0.56	&\textbf{0.18}	&0.33	&0.42	&0.39	\\
					&KPOD	&0.63	&0.18	&0.49	&0.06	&0.51	&0.14	&0.29	&0.37	&0.64	&0.25	&0.16	&0.33	&0.41	&0.34	\\
					&TDM	&0.65	&0.45	&0.33	&0.05	&-0.08	&0.16	&0.41	&0.48	&0.8	&0.52	&0.17	&0.32	&0.38	&0.36	\\
					&HyperImpute	&0.73	&0.49	&0.38	&0.05	&-0.08	&0.17	&0.41	&\textbf{0.63}	&0.8	&0.52	&\textbf{0.18}	&0.33	&0.41	&0.39	\\
					&MBFS	&0.73	&0.49	&0.38	&0.05	&-0.08	&0.17	&\textbf{0.49}	&0.6	&\textbf{0.82}	&\textbf{0.6}	&\textbf{0.18}	&0.33	&-\tnote{1}	&0.37	\\
					&DIMV	&0.69	&0.48	&0.36	&0.05	&-0.08	&0.14	&0.43	&0.48	&0.76	&0.52	&0.17	&0.34	&0.44	&0.37	\\
					
					\bottomrule
					
					\toprule
					\multicolumn{2}{c}{Dataset} & \emph{Iris} & \emph{Brea} & \emph{Wine} & \emph{Bank} & \emph{Htru} & \emph{Know} & \emph{Bir1} & \emph{Bir2} & \emph{Ove1} & \emph{Ove2} & \emph{Worm} & \emph{Urba} & \emph{Winn} & \textbf{Avg}\\
					\midrule
					\multirow{11}{*}{NMI}
					&\textbf{SDC (Ours)}	&\textbf{0.8}	&\textbf{0.49}	&\textbf{0.59}	&\textbf{0.25}	&\textbf{0.39}	&\textbf{0.34}	&0.8	&\textbf{0.87}	&0.65	&\textbf{0.74}	&\textbf{0.46}	&\textbf{0.51}	&0.6	&\textbf{0.58}	\\
					&MICE	&0.75	&0.46	&0.43	&0.03	&0.03	&0.19	&0.76	&0.81	&0.7	&0.71	&0.42	&0.5	&0.59	&0.49	\\
					&LRC	&0.48	&0.45	&0.35	&0.03	&0.03	&0.16	&0.76	&0.81	&0.61	&0.7	&0.41	&0.38	&0.49	&0.44	\\
					&GAIN	&0.69	&0.46	&0.39	&0.03	&0.01	&0.21	&\textbf{0.81}	&0.81	&0.62	&0.7	&0.42	&0.5	&\textbf{0.64}	&0.48	\\
					&MDIOT	&0.69	&0.43	&0.39	&0.03	&0.02	&0.2	&\textbf{0.81}	&0.81	&0.7	&0.71	&0.42	&0.5	&0.63	&0.49	\\
					&VAEAC	&0.76	&0.46	&0.41	&0.03	&0.03	&0.18	&0.75	&0.84	&0.71	&0.71	&0.43	&0.5	&\textbf{0.64}	&0.5	\\
					&KPOD	&0.68	&0.21	&0.56	&0.07	&\textbf{0.39}	&0.19	&0.69	&0.76	&0.62	&0.49	&0.43	&0.46	&0.54	&0.47	\\
					&TDM	&0.67	&0.43	&0.4	&0.03	&0.02	&0.2	&0.76	&0.81	&0.7	&0.71	&0.42	&0.48	&0.61	&0.48	\\
					&HyperImpute	&0.75	&0.46	&0.44	&0.03	&0.03	&0.21	&0.75	&0.82	&0.71	&0.71	&0.42	&0.5	&0.63	&0.5	\\
					&MBFS	&0.76	&0.46	&0.44	&0.03	&0.03	&0.2	&0.69	&0.76	&\textbf{0.72}	&0.67	&0.41	&0.5	&-\tnote{1}	&0.44	\\
					&DIMV	&0.71	&0.46	&0.42	&0.03	&0.03	&0.18	&0.77	&0.81	&0.67	&0.71	&0.42	&0.48	&0.62	&0.49	\\
					
					\bottomrule
					
					\toprule
					\multicolumn{2}{c}{Dataset} & \emph{Iris} & \emph{Brea} & \emph{Wine} & \emph{Bank} & \emph{Htru} & \emph{Know} & \emph{Bir1} & \emph{Bir2} & \emph{Ove1} & \emph{Ove2} & \emph{Worm} & \emph{Urba} & \emph{Winn} & \textbf{Avg}\\
					\midrule
					\multirow{11}{*}{PUR}
					&\textbf{SDC (Ours)}	&\textbf{0.92}	&\textbf{0.91}	&\textbf{0.89}	&\textbf{0.83}	&\textbf{0.96}	&\textbf{0.64}	&\textbf{0.72}	&\textbf{0.85}	&\textbf{0.95}	&\textbf{0.81}	&\textbf{0.39}	&0.54	&0.81	&\textbf{0.79}	\\
					&MICE	&0.88	&0.85	&0.69	&0.62	&0.91	&0.53	&0.63	&0.56	&\textbf{0.95}	&0.75	&0.29	&\textbf{0.69}	&0.82	&0.71	\\
					&LRC	&0.71	&0.85	&0.67	&0.6	&0.91	&0.49	&0.64	&0.56	&0.9	&0.74	&0.29	&0.58	&0.74	&0.67	\\
					&GAIN	&0.87	&0.85	&0.67	&0.6	&0.91	&0.54	&0.58	&0.58	&0.92	&0.75	&0.29	&\textbf{0.69}	&\textbf{0.87}	&0.7	\\
					&MDIOT	&0.87	&0.84	&0.67	&0.61	&0.91	&0.53	&0.58	&0.58	&0.94	&0.75	&0.29	&\textbf{0.69}	&0.86	&0.7	\\
					&VAEAC	&0.9	&0.85	&0.68	&0.61	&0.91	&0.52	&0.66	&0.69	&\textbf{0.95}	&0.79	&0.3	&\textbf{0.69}	&0.86	&0.72	\\
					&KPOD	&0.84	&0.73	&0.74	&0.62	&0.95	&0.52	&0.51	&0.46	&0.9	&0.43	&0.3	&0.62	&0.7	&0.64	\\
					&TDM	&0.86	&0.84	&0.69	&0.61	&0.91	&0.54	&0.63	&0.58	&\textbf{0.95}	&0.75	&0.29	&0.67	&0.84	&0.7	\\
					&HyperImpute	&0.89	&0.85	&0.71	&0.61	&0.91	&0.54	&0.63	&0.68	&\textbf{0.95}	&0.74	&0.3	&\textbf{0.69}	&\textbf{0.87}	&0.72	\\
					&MBFS	&0.89	&0.85	&0.71	&0.61	&0.91	&0.54	&0.69	&0.62	&\textbf{0.95}	&0.8	&0.29	&\textbf{0.69}	&-\tnote{1}	&0.66	\\
					&DIMV	&0.87	&0.85	&0.7	&0.62	&0.91	&0.52	&0.66	&0.58	&0.93	&0.75	&0.29	&0.65	&0.82	&0.7	\\
					
					\bottomrule
				\end{tabular}
			\end{sc}
		\end{small}
		\begin{tablenotes}
			\footnotesize
			\item[1] Runtime is too long to get results.
		\end{tablenotes}
	\end{center}
	\end{threeparttable}
	\vskip -0.1in
\end{table*}

\begin{table*}
	\caption{The accuracies of SDC and baseline algorithms on 13 missing datasets with a missing rate of 0.4}
	\label{table5}
	\vskip 0.15in
	\setlength{\tabcolsep}{4.7pt}
	\begin{threeparttable}
	\begin{center}
		\begin{small}
			\begin{sc}
				\begin{tabular}{c|cccccccccccccc|c}
					\toprule
					\multicolumn{2}{c}{Dataset} & \emph{Iris} & \emph{Brea} & \emph{Wine} & \emph{Bank} & \emph{Htru} & \emph{Know} & \emph{Bir1} & \emph{Bir2} & \emph{Ove1} & \emph{Ove2} & \emph{Worm} & \emph{Urba} & \emph{Winn} & \textbf{Avg}\\
					\midrule
					\multirow{11}{*}{ARI}
					&\textbf{SDC (Ours)}	&\textbf{0.77}	&\textbf{0.63}	&\textbf{0.56}	&\textbf{0.15}	&\textbf{0.52}	&\textbf{0.26}	&0.3	&0.21	&0.74	&0.46	&\textbf{0.16}	&\textbf{0.41}	&\textbf{0.57}	&\textbf{0.44}	\\
					&MICE	&0.71	&0.49	&0.37	&0.05	&-0.08	&0.15	&0.31	&0.43	&0.78	&0.45	&0.15	&0.33	&0.36	&0.35	\\
					&LRC	&0.42	&0.46	&0.31	&0.04	&-0.08	&0.1	&0.32	&0.4	&0.53	&0.4	&0.15	&0.24	&0.31	&0.28	\\
					&GAIN	&0.66	&0.49	&0.31	&0.03	&-0.05	&0.17	&\textbf{0.39}	&0.39	&0.6	&0.46	&-0.33	&0.33	&0.41	&0.3	\\
					&MDIOT	&0.65	&0.43	&0.3	&0.05	&-0.07	&0.15	&0.38	&0.38	&0.76	&0.45	&0.15	&0.33	&0.41	&0.34	\\
					&VAEAC	&0.75	&0.49	&0.35	&0.05	&-0.08	&0.17	&0.34	&\textbf{0.56}	&0.79	&0.47	&0.15	&0.33	&0.41	&0.37	\\
					&KPOD	&0.59	&0.16	&0.46	&0.05	&0.49	&0.13	&0.22	&0.29	&0.58	&0.21	&0.13	&0.31	&0.41	&0.31	\\
					&TDM	&0.63	&0.43	&0.31	&0.05	&-0.08	&0.14	&0.31	&0.38	&0.78	&0.45	&0.15	&0.32	&0.41	&0.33	\\
					&HyperImpute	&0.7	&0.49	&0.37	&0.03	&-0.08	&0.14	&0.31	&0.54	&0.79	&0.47	&\textbf{0.16}	&0.33	&0.41	&0.36	\\
					&MBFS	&0.72	&0.49	&0.35	&0.05	&-0.08	&0.17	&0.37	&0.52	&\textbf{0.81}	&\textbf{0.5}	&\textbf{0.16}	&0.39	&-\tnote{1}	&0.34	\\
					&DIMV	&0.66	&0.46	&0.36	&0.05	&-0.08	&0.15	&0.32	&0.39	&0.72	&0.43	&0.15	&0.33	&0.42	&0.34	\\
					
					\bottomrule
					
					\toprule
					\multicolumn{2}{c}{Dataset} & \emph{Iris} & \emph{Brea} & \emph{Wine} & \emph{Bank} & \emph{Htru} & \emph{Know} & \emph{Bir1} & \emph{Bir2} & \emph{Ove1} & \emph{Ove2} & \emph{Worm} & \emph{Urba} & \emph{Winn} & \textbf{Avg}\\
					\midrule
					\multirow{11}{*}{NMI}
					&\textbf{SDC (Ours)}	&\textbf{0.79}	&\textbf{0.48}	&\textbf{0.57}	&\textbf{0.24}	&\textbf{0.37}	&\textbf{0.32}	&0.75	&\textbf{0.83}	&0.62	&\textbf{0.7}	&\textbf{0.44}	&0.5	&0.6	&\textbf{0.55}	\\
					&MICE	&0.74	&0.46	&0.43	&0.03	&0.03	&0.19	&0.71	&0.77	&0.68	&0.67	&0.4	&0.51	&0.59	&0.48	\\
					&LRC	&0.42	&0.44	&0.33	&0.02	&0.03	&0.14	&0.71	&0.77	&0.51	&0.66	&0.38	&0.37	&0.48	&0.4	\\
					&GAIN	&0.69	&0.46	&0.38	&0.02	&0.01	&0.22	&\textbf{0.77}	&0.77	&0.55	&0.66	&0.4	&0.5	&\textbf{0.64}	&0.47	\\
					&MDIOT	&0.67	&0.42	&0.37	&0.03	&0.02	&0.19	&\textbf{0.77}	&0.77	&0.67	&0.67	&0.4	&0.5	&0.63	&0.47	\\
					&VAEAC	&0.77	&0.46	&0.41	&0.03	&0.03	&0.22	&0.7	&0.82	&0.69	&0.66	&0.39	&0.5	&0.63	&0.49	\\
					&KPOD	&0.66	&0.2	&0.53	&0.06	&\textbf{0.37}	&0.18	&0.65	&0.73	&0.57	&0.46	&0.4	&0.45	&0.53	&0.45	\\
					&TDM	&0.65	&0.42	&0.39	&0.03	&0.02	&0.17	&0.71	&0.77	&0.69	&0.67	&0.4	&0.48	&0.63	&0.46	\\
					&HyperImpute	&0.74	&0.46	&0.43	&0.02	&0.03	&0.17	&0.7	&0.77	&0.69	&0.62	&0.4	&0.5	&0.63	&0.47	\\
					&MBFS	&0.75	&0.46	&0.42	&0.03	&0.03	&0.22	&0.62	&0.7	&\textbf{0.71}	&0.6	&0.35	&\textbf{0.53}	&-\tnote{1}	&0.42	\\
					&DIMV	&0.7	&0.43	&0.42	&0.03	&0.03	&0.19	&0.71	&0.77	&0.64	&0.66	&0.4	&0.47	&0.58	&0.46	\\
					
					\bottomrule
					
					\toprule
					\multicolumn{2}{c}{Dataset} & \emph{Iris} & \emph{Brea} & \emph{Wine} & \emph{Bank} & \emph{Htru} & \emph{Know} & \emph{Bir1} & \emph{Bir2} & \emph{Ove1} & \emph{Ove2} & \emph{Worm} & \emph{Urba} & \emph{Winn} & \textbf{Avg}\\
					\midrule
					\multirow{11}{*}{PUR}
					&\textbf{SDC (Ours)}	&\textbf{0.91}	&\textbf{0.91}	&\textbf{0.89}	&\textbf{0.81}	&\textbf{0.96}	&\textbf{0.63}	&\textbf{0.63}	&\textbf{0.8}	&0.94	&\textbf{0.77}	&\textbf{0.39}	&0.53	&0.8	&\textbf{0.77}	\\
					&MICE	&0.88	&0.85	&0.69	&0.62	&0.91	&0.53	&0.54	&0.5	&0.94	&0.7	&0.27	&0.69	&0.82	&0.69	\\
					&LRC	&0.69	&0.84	&0.65	&0.6	&0.91	&0.48	&0.55	&0.5	&0.86	&0.69	&0.27	&0.57	&0.73	&0.64	\\
					&GAIN	&0.86	&0.85	&0.66	&0.59	&0.91	&0.54	&0.51	&0.51	&0.88	&0.7	&0.27	&0.69	&\textbf{0.87}	&0.68	\\
					&MDIOT	&0.86	&0.83	&0.66	&0.61	&0.91	&0.54	&0.51	&0.51	&0.94	&0.7	&0.26	&0.69	&0.86	&0.68	\\
					&VAEAC	&0.9	&0.85	&0.68	&0.61	&0.91	&0.56	&0.57	&0.68	&0.94	&0.74	&0.27	&0.69	&0.86	&0.71	\\
					&KPOD	&0.82	&0.72	&0.73	&0.61	&0.94	&0.52	&0.45	&0.39	&0.88	&0.4	&0.28	&0.61	&0.7	&0.62	\\
					&TDM	&0.85	&0.83	&0.68	&0.61	&0.91	&0.52	&0.54	&0.52	&0.94	&0.7	&0.26	&0.66	&0.85	&0.68	\\
					&HyperImpute	&0.88	&0.85	&0.71	&0.6	&0.91	&0.51	&0.54	&0.6	&0.94	&0.71	&0.28	&0.69	&0.86	&0.7	\\
					&MBFS	&0.89	&0.85	&0.69	&0.61	&0.91	&0.53	&0.6	&0.54	&\textbf{0.95}	&0.73	&0.28	&\textbf{0.7}	&-\tnote{1}	&0.64	\\
					&DIMV	&0.86	&0.84	&0.7	&0.62	&0.91	&0.53	&0.56	&0.51	&0.92	&0.69	&0.27	&0.62	&0.78	&0.68	\\
					
					\bottomrule
				\end{tabular}
			\end{sc}
		\end{small}
		\begin{tablenotes}
			\footnotesize
			\item[1] Runtime is too long to get results.
		\end{tablenotes}
	\end{center}
	\end{threeparttable}
	\vskip -0.1in
\end{table*}

	\begin{table*}
	\caption{The accuracies of SDC and baseline algorithms on 13 missing datasets with a missing rate of 0.5}
	\label{table6}
	\vskip 0.15in
	\setlength{\tabcolsep}{4.7pt}
	\begin{threeparttable}
	\begin{center}
		\begin{small}
			\begin{sc}
				\begin{tabular}{c|cccccccccccccc|c}
					\toprule
					\multicolumn{2}{c}{Dataset} & \emph{Iris} & \emph{Brea} & \emph{Wine} & \emph{Bank} & \emph{Htru} & \emph{Know} & \emph{Bir1} & \emph{Bir2} & \emph{Ove1} & \emph{Ove2} & \emph{Worm} & \emph{Urba} & \emph{Winn} & \textbf{Avg}\\
					\midrule
					\multirow{11}{*}{ARI}
					&\textbf{SDC (Ours)}	&0.65	&\textbf{0.62}	&\textbf{0.52}	&\textbf{0.15}	&\textbf{0.55}	&\textbf{0.33}	&0.24	&0.13	&0.73	&0.33	&0.13	&\textbf{0.4}	&\textbf{0.57}	&\textbf{0.41}	\\
					&MICE	&0.7	&0.49	&0.36	&0.06	&-0.08	&0.14	&0.23	&0.37	&0.77	&0.38	&0.13	&0.33	&0.36	&0.33	\\
					&LRC	&0.39	&0.44	&0.3	&0.04	&-0.08	&0.09	&0.24	&0.33	&0.56	&0.32	&0.13	&0.23	&0.3	&0.25	\\
					&GAIN	&0.65	&0.49	&0.31	&0.04	&-0.04	&0.16	&\textbf{0.33}	&0.33	&0.52	&0.39	&-0.7	&0.33	&0.41	&0.25	\\
					&MDIOT	&0.64	&0.41	&0.28	&0.05	&-0.07	&0.13	&0.31	&0.31	&0.52	&0.39	&0.13	&0.33	&0.41	&0.3	\\
					&VAEAC	&\textbf{0.74}	&0.49	&0.35	&0.05	&-0.08	&0.15	&0.25	&\textbf{0.5}	&0.77	&\textbf{0.4}	&\textbf{0.14}	&0.33	&0.41	&0.35	\\
					&KPOD	&0.56	&0.14	&0.47	&0.06	&0.5	&0.13	&0.17	&0.22	&0.51	&0.19	&0.11	&0.32	&0.41	&0.29	\\
					&TDM	&0.62	&0.41	&0.3	&0.05	&-0.08	&0.14	&0.23	&0.31	&0.54	&0.39	&0.13	&0.33	&0.4	&0.29	\\
					&HyperImpute	&0.7	&0.49	&0.39	&0.04	&-0.08	&0.13	&0.23	&0.47	&0.77	&0.39	&0.13	&0.33	&0.42	&0.34	\\
					&MBFS	&0.73	&0.49	&0.39	&0.05	&-0.08	&0.13	&0.28	&0.44	&\textbf{0.8}	&0.39	&0.13	&0.33	&-\tnote{1}	&0.31	\\
					&DIMV	&0.68	&0.35	&0.35	&0.05	&-0.08	&0.15	&0.24	&0.31	&0.69	&0.37	&0.13	&0.32	&0.39	&0.3	\\
					
					\bottomrule
					
					\toprule
					\multicolumn{2}{c}{Dataset} & \emph{Iris} & \emph{Brea} & \emph{Wine} & \emph{Bank} & \emph{Htru} & \emph{Know} & \emph{Bir1} & \emph{Bir2} & \emph{Ove1} & \emph{Ove2} & \emph{Worm} & \emph{Urba} & \emph{Winn} & \textbf{Avg}\\
					\midrule
					\multirow{11}{*}{NMI}
					&\textbf{SDC (Ours)}	&0.75	&\textbf{0.47}	&0.52	&\textbf{0.24}	&\textbf{0.39}	&\textbf{0.37}	&0.72	&\textbf{0.79}	&0.6	&\textbf{0.63}	&\textbf{0.41}	&0.5	&0.59	&\textbf{0.54}	\\
					&MICE	&0.74	&0.46	&0.42	&0.03	&0.03	&0.18	&0.66	&0.75	&0.66	&\textbf{0.63}	&0.37	&\textbf{0.51}	&0.59	&0.46	\\
					&LRC	&0.4	&0.43	&0.32	&0.02	&0.03	&0.13	&0.66	&0.75	&0.54	&0.61	&0.35	&0.36	&0.47	&0.39	\\
					&GAIN	&0.67	&0.45	&0.38	&0.02	&0.01	&0.21	&\textbf{0.75}	&0.75	&0.49	&0.61	&0.37	&0.5	&\textbf{0.64}	&0.45	\\
					&MDIOT	&0.65	&0.4	&0.36	&0.03	&0.02	&0.17	&\textbf{0.75}	&0.75	&0.51	&\textbf{0.63}	&0.37	&0.5	&0.63	&0.44	\\
					&VAEAC	&\textbf{0.77}	&\textbf{0.47}	&0.41	&0.03	&0.03	&0.19	&0.65	&0.75	&0.66	&0.62	&0.37	&0.5	&0.63	&0.47	\\
					&KPOD	&0.62	&0.18	&\textbf{0.54}	&0.07	&0.37	&0.18	&0.62	&0.71	&0.52	&0.43	&0.37	&0.45	&0.53	&0.43	\\
					&TDM	&0.64	&0.4	&0.38	&0.03	&0.02	&0.17	&0.66	&0.75	&0.52	&\textbf{0.63}	&0.37	&0.48	&0.61	&0.44	\\
					&HyperImpute	&0.74	&0.46	&0.44	&0.02	&0.03	&0.17	&0.65	&0.73	&0.67	&0.55	&0.37	&0.5	&\textbf{0.64}	&0.46	\\
					&MBFS	&0.76	&0.46	&0.44	&0.03	&0.03	&0.15	&0.56	&0.64	&\textbf{0.7}	&0.5	&0.31	&0.5	&-\tnote{1}	&0.39	\\
					&DIMV	&0.7	&0.34	&0.41	&0.03	&0.03	&0.19	&0.67	&0.75	&0.61	&0.62	&0.37	&0.45	&0.55	&0.44	\\
					
					\bottomrule
					
					\toprule
					\multicolumn{2}{c}{Dataset} & \emph{Iris} & \emph{Brea} & \emph{Wine} & \emph{Bank} & \emph{Htru} & \emph{Know} & \emph{Bir1} & \emph{Bir2} & \emph{Ove1} & \emph{Ove2} & \emph{Worm} & \emph{Urba} & \emph{Winn} & \textbf{Avg}\\
					\midrule
					\multirow{11}{*}{PUR}
					&\textbf{SDC (Ours)}	&0.85	&\textbf{0.91}	&\textbf{0.87}	&\textbf{0.81}	&\textbf{0.96}	&\textbf{0.73}	&\textbf{0.54}	&\textbf{0.75}	&0.93	&\textbf{0.69}	&\textbf{0.36}	&0.55	&0.79	&\textbf{0.75}	\\
					&MICE	&0.88	&0.85	&0.68	&0.62	&0.91	&0.52	&0.45	&0.45	&0.94	&0.65	&0.25	&\textbf{0.69}	&0.82	&0.67	\\
					&LRC	&0.67	&0.84	&0.65	&0.6	&0.91	&0.47	&0.46	&0.45	&0.87	&0.64	&0.25	&0.56	&0.72	&0.62	\\
					&GAIN	&0.86	&0.85	&0.66	&0.59	&0.91	&0.53	&0.46	&0.46	&0.85	&0.65	&0.25	&\textbf{0.69}	&0.86	&0.66	\\
					&MDIOT	&0.86	&0.83	&0.66	&0.61	&0.91	&0.51	&0.46	&0.46	&0.85	&0.66	&0.24	&\textbf{0.69}	&0.86	&0.66	\\
					&VAEAC	&\textbf{0.9}	&0.85	&0.69	&0.62	&0.91	&0.52	&0.48	&0.57	&0.94	&0.66	&0.26	&\textbf{0.69}	&0.86	&0.69	\\
					&KPOD	&0.8	&0.71	&0.73	&0.62	&0.94	&0.51	&0.38	&0.4	&0.85	&0.37	&0.26	&0.61	&0.7	&0.61	\\
					&TDM	&0.84	&0.83	&0.67	&0.61	&0.91	&0.52	&0.45	&0.46	&0.86	&0.65	&0.24	&0.66	&0.85	&0.66	\\
					&HyperImpute	&0.88	&0.85	&0.72	&0.6	&0.91	&0.51	&0.46	&0.54	&0.94	&0.63	&0.25	&\textbf{0.69}	&\textbf{0.87}	&0.68	\\
					&MBFS	&0.89	&0.85	&0.71	&0.61	&0.91	&0.51	&0.52	&0.46	&\textbf{0.95}	&0.63	&0.25	&\textbf{0.69}	&-\tnote{1}	&0.61	\\
					&DIMV	&0.87	&0.79	&0.69	&0.62	&0.91	&0.52	&0.47	&0.46	&0.91	&0.64	&0.24	&0.6	&0.74	&0.65	\\
					
					\bottomrule
				\end{tabular}
			\end{sc}
		\end{small}
		\begin{tablenotes}
			\footnotesize
			\item[1] Runtime is too long to get results.
		\end{tablenotes}
	\end{center}
	\end{threeparttable}
	\vskip -0.1in
\end{table*}

	\begin{table*}
	\caption{The accuracies of SDC and baseline algorithms on 13 missing datasets with a missing rate of 0.6}
	\label{table7}
	\vskip 0.15in
	\setlength{\tabcolsep}{4.7pt}
	\begin{threeparttable}
	\begin{center}
		\begin{small}
			\begin{sc}
				\begin{tabular}{c|cccccccccccccc|c}
					\toprule
					\multicolumn{2}{c}{Dataset} & \emph{Iris} & \emph{Brea} & \emph{Wine} & \emph{Bank} & \emph{Htru} & \emph{Know} & \emph{Bir1} & \emph{Bir2} & \emph{Ove1} & \emph{Ove2} & \emph{Worm} & \emph{Urba} & \emph{Winn} & \textbf{Avg}\\
					\midrule
					\multirow{11}{*}{ARI}
					&\textbf{SDC (Ours)}	&0.65	&\textbf{0.61}	&\textbf{0.52}	&\textbf{0.15}	&\textbf{0.54}	&\textbf{0.31}	&0.17	&0.09	&0.69	&0.26	&0.11	&\textbf{0.4}	&\textbf{0.56}	&\textbf{0.39}	\\
					&MICE	&0.71	&0.49	&0.38	&0.06	&-0.08	&0.15	&0.18	&0.31	&0.74	&0.34	&0.11	&0.33	&0.36	&0.31	\\
					&LRC	&0.36	&0.43	&0.28	&0.04	&-0.08	&0.09	&0.19	&0.27	&0.49	&0.3	&0.11	&0.22	&0.29	&0.23	\\
					&GAIN	&0.66	&0.49	&0.32	&0.04	&-0.04	&0.16	&\textbf{0.27}	&0.27	&0.44	&0.33	&-0.2	&0.33	&0.39	&0.27	\\
					&MDIOT	&0.63	&0.4	&0.28	&0.05	&-0.07	&0.15	&0.26	&0.26	&0.39	&0.33	&0.11	&0.33	&0.41	&0.27	\\
					&VAEAC	&\textbf{0.76}	&0.49	&0.36	&0.05	&-0.08	&0.18	&0.19	&\textbf{0.38}	&0.68	&0.33	&\textbf{0.12}	&0.33	&0.39	&0.32	\\
					&KPOD	&0.55	&0.13	&0.45	&0.06	&0.48	&0.13	&0.14	&0.17	&0.41	&0.17	&0.09	&0.31	&0.41	&0.27	\\
					&TDM	&0.58	&0.4	&0.28	&0.05	&-0.07	&0.13	&0.18	&0.25	&0.44	&0.33	&0.11	&0.33	&0.39	&0.26	\\
					&HyperImpute	&0.7	&0.5	&0.39	&0.04	&-0.08	&0.19	&0.18	&0.37	&\textbf{0.76}	&0.34	&\textbf{0.12}	&0.33	&0.4	&0.33	\\
					&MBFS	&0.71	&0.49	&0.41	&0.05	&-0.08	&0.15	&0.2	&0.37	&0.75	&\textbf{0.35}	&0.11	&0.33	&-\tnote{1}	&0.3	\\
					&DIMV	&0.66	&0.41	&0.34	&0.05	&-0.08	&0.16	&0.18	&0.26	&0.63	&0.31	&0.11	&0.29	&0.39	&0.29	\\
					
					\bottomrule
					
					\toprule
					\multicolumn{2}{c}{Dataset} & \emph{Iris} & \emph{Brea} & \emph{Wine} & \emph{Bank} & \emph{Htru} & \emph{Know} & \emph{Bir1} & \emph{Bir2} & \emph{Ove1} & \emph{Ove2} & \emph{Worm} & \emph{Urba} & \emph{Winn} & \textbf{Avg}\\
					\midrule
					\multirow{11}{*}{NMI}
					&\textbf{SDC (Ours)}	&0.75	&0.46	&\textbf{0.53}	&\textbf{0.23}	&\textbf{0.38}	&\textbf{0.34}	&0.67	&\textbf{0.75}	&0.56	&0.59	&\textbf{0.39}	&0.49	&0.59	&\textbf{0.52}	\\
					&MICE	&0.74	&0.46	&0.44	&0.03	&0.03	&0.2	&0.62	&0.72	&0.63	&0.57	&0.35	&0.5	&0.59	&0.45	\\
					&LRC	&0.37	&0.42	&0.29	&0.02	&0.03	&0.12	&0.62	&0.73	&0.47	&0.59	&0.33	&0.35	&0.46	&0.37	\\
					&GAIN	&0.68	&0.46	&0.38	&0.02	&0.01	&0.22	&\textbf{0.73}	&0.73	&0.43	&0.58	&0.35	&0.5	&0.62	&0.44	\\
					&MDIOT	&0.65	&0.39	&0.36	&0.03	&0.02	&0.19	&\textbf{0.73}	&0.73	&0.42	&\textbf{0.6}	&0.35	&0.5	&\textbf{0.63}	&0.43	\\
					&VAEAC	&\textbf{0.77}	&0.46	&0.42	&0.03	&0.03	&0.23	&0.6	&0.71	&0.59	&0.57	&0.35	&0.5	&0.61	&0.45	\\
					&KPOD	&0.63	&0.16	&0.52	&0.07	&0.36	&0.19	&0.59	&0.69	&0.45	&0.41	&0.35	&0.45	&0.53	&0.42	\\
					&TDM	&0.61	&0.39	&0.36	&0.03	&0.02	&0.17	&0.62	&0.73	&0.45	&\textbf{0.6}	&0.35	&0.48	&0.58	&0.41	\\
					&HyperImpute	&0.74	&\textbf{0.47}	&0.44	&0.02	&0.03	&0.24	&0.61	&0.7	&\textbf{0.66}	&0.51	&0.33	&\textbf{0.51}	&\textbf{0.63}	&0.45	\\
					&MBFS	&0.74	&0.46	&0.44	&0.03	&0.03	&0.21	&0.5	&0.58	&0.65	&0.46	&0.28	&0.5	&-\tnote{1}	&0.38	\\
					&DIMV	&0.69	&0.38	&0.4	&0.03	&0.03	&0.2	&0.63	&0.73	&0.57	&0.59	&0.35	&0.41	&0.54	&0.43	\\
					
					\bottomrule
					
					\toprule
					\multicolumn{2}{c}{Dataset} & \emph{Iris} & \emph{Brea} & \emph{Wine} & \emph{Bank} & \emph{Htru} & \emph{Know} & \emph{Bir1} & \emph{Bir2} & \emph{Ove1} & \emph{Ove2} & \emph{Worm} & \emph{Urba} & \emph{Winn} & \textbf{Avg}\\
					\midrule
					\multirow{11}{*}{PUR}
					&\textbf{SDC (Ours)}	&0.85	&\textbf{0.9}	&\textbf{0.87}	&\textbf{0.81}	&\textbf{0.96}	&\textbf{0.72}	&\textbf{0.45}	&\textbf{0.7}	&0.92	&\textbf{0.65}	&\textbf{0.28}	&0.55	&0.79	&\textbf{0.73}	\\
					&MICE	&0.88	&0.85	&0.7	&0.62	&0.91	&0.52	&0.37	&0.42	&\textbf{0.93}	&0.6	&0.23	&0.68	&0.82	&0.66	\\
					&LRC	&0.66	&0.83	&0.64	&0.6	&0.91	&0.47	&0.37	&0.41	&0.84	&0.59	&0.23	&0.54	&0.71	&0.6	\\
					&GAIN	&0.86	&0.85	&0.67	&0.6	&0.91	&0.53	&0.42	&0.42	&0.82	&0.6	&0.23	&\textbf{0.69}	&0.85	&0.65	\\
					&MDIOT	&0.86	&0.82	&0.66	&0.61	&0.91	&0.52	&0.42	&0.42	&0.81	&0.61	&0.23	&\textbf{0.69}	&\textbf{0.86}	&0.65	\\
					&VAEAC	&\textbf{0.91}	&0.85	&0.69	&0.61	&0.91	&0.55	&0.39	&0.49	&0.91	&0.61	&0.23	&\textbf{0.69}	&0.84	&0.67	\\
					&KPOD	&0.8	&0.7	&0.71	&0.62	&0.94	&0.51	&0.33	&0.36	&0.82	&0.34	&0.23	&0.61	&0.7	&0.59	\\
					&TDM	&0.83	&0.82	&0.66	&0.61	&0.91	&0.51	&0.37	&0.42	&0.82	&0.61	&0.22	&0.65	&0.82	&0.63	\\
					&HyperImpute	&0.88	&0.86	&0.71	&0.6	&0.91	&0.55	&0.38	&0.47	&\textbf{0.93}	&0.6	&0.23	&0.68	&0.85	&0.67	\\
					&MBFS	&0.88	&0.85	&0.74	&0.61	&0.91	&0.53	&0.42	&0.4	&\textbf{0.93}	&0.61	&0.23	&0.68	&-\tnote{1}	&0.6	\\
					&DIMV	&0.86	&0.82	&0.69	&0.61	&0.91	&0.53	&0.38	&0.42	&0.89	&0.6	&0.23	&0.57	&0.73	&0.63	\\
					
					\bottomrule
				\end{tabular}
			\end{sc}
		\end{small}
		\begin{tablenotes}
			\footnotesize
			\item[1] Runtime is too long to get results.
		\end{tablenotes}
	\end{center}
	\end{threeparttable}
	\vskip -0.1in
\end{table*}

	\begin{table*}
	\caption{The accuracies of SDC and baseline algorithms on 13 missing datasets with a missing rate of 0.7}
	\label{table8}
	\vskip 0.15in
	\setlength{\tabcolsep}{4.7pt}
	\begin{threeparttable}
	\begin{center}
		\begin{small}
			\begin{sc}
				\begin{tabular}{c|cccccccccccccc|c}
					\toprule
					\multicolumn{2}{c}{Dataset} & \emph{Iris} & \emph{Brea} & \emph{Wine} & \emph{Bank} & \emph{Htru} & \emph{Know} & \emph{Bir1} & \emph{Bir2} & \emph{Ove1} & \emph{Ove2} & \emph{Worm} & \emph{Urba} & \emph{Winn} & \textbf{Avg}\\
					\midrule
					\multirow{11}{*}{ARI}
					&\textbf{SDC (Ours)}	&0.68	&\textbf{0.61}	&\textbf{0.52}	&\textbf{0.14}	&\textbf{0.55}	&\textbf{0.28}	&0.14	&0.06	&0.7	&0.19	&\textbf{0.1}	&\textbf{0.4}	&\textbf{0.57}	&\textbf{0.38}	\\
					&MICE	&0.71	&0.48	&0.37	&0.06	&-0.08	&0.15	&0.15	&0.28	&\textbf{0.75}	&\textbf{0.29}	&\textbf{0.1}	&0.33	&0.36	&0.3	\\
					&LRC	&0.34	&0.42	&0.26	&0.03	&-0.08	&0.08	&0.15	&0.24	&0.46	&0.26	&\textbf{0.1}	&0.21	&0.28	&0.21	\\
					&GAIN	&0.63	&0.49	&0.32	&0.04	&-0.04	&0.15	&\textbf{0.24}	&0.24	&0.41	&0.27	&-0.73	&0.33	&0.41	&0.21	\\
					&MDIOT	&0.6	&0.38	&0.26	&0.04	&-0.07	&0.14	&0.22	&0.22	&0.32	&0.28	&\textbf{0.1}	&0.33	&0.41	&0.25	\\
					&VAEAC	&\textbf{0.76}	&0.5	&0.36	&0.05	&-0.08	&0.16	&0.14	&0.27	&\textbf{0.75}	&0.27	&\textbf{0.1}	&0.32	&0.42	&0.31	\\
					&KPOD	&0.52	&0.14	&0.41	&0.05	&0.47	&0.14	&0.12	&0.15	&0.3	&0.15	&0.09	&0.3	&0.41	&0.25	\\
					&TDM	&0.54	&0.38	&0.24	&0.04	&-0.07	&0.14	&0.15	&0.22	&0.35	&0.28	&\textbf{0.1}	&0.33	&0.41	&0.24	\\
					&HyperImpute	&0.7	&0.5	&0.34	&0.03	&-0.08	&0.12	&0.15	&\textbf{0.32}	&0.74	&0.28	&\textbf{0.1}	&0.34	&0.42	&0.3	\\
					&MBFS	&0.72	&0.49	&0.38	&0.04	&-0.08	&0.17	&0.14	&0.3	&\textbf{0.75}	&\textbf{0.29}	&0.09	&0.33	&-\tnote{1}	&0.28	\\
					&DIMV	&0.63	&0.38	&0.32	&0.05	&-0.08	&0.16	&0.15	&0.24	&0.63	&0.28	&\textbf{0.1}	&0.28	&0.34	&0.27	\\
					
					\bottomrule
					
					\toprule
					\multicolumn{2}{c}{Dataset} & \emph{Iris} & \emph{Brea} & \emph{Wine} & \emph{Bank} & \emph{Htru} & \emph{Know} & \emph{Bir1} & \emph{Bir2} & \emph{Ove1} & \emph{Ove2} & \emph{Worm} & \emph{Urba} & \emph{Winn} & \textbf{Avg}\\
					\midrule
					\multirow{11}{*}{NMI}
					&\textbf{SDC (Ours)}	&0.74	&0.45	&\textbf{0.53}	&\textbf{0.22}	&\textbf{0.39}	&\textbf{0.32}	&0.64	&\textbf{0.72}	&0.57	&0.55	&\textbf{0.37}	&0.48	&0.58	&\textbf{0.5}	\\
					&MICE	&0.75	&0.46	&0.43	&0.04	&0.03	&0.19	&0.59	&0.69	&\textbf{0.65}	&0.47	&0.33	&\textbf{0.5}	&0.59	&0.44	\\
					&LRC	&0.35	&0.41	&0.28	&0.02	&0.03	&0.11	&0.58	&0.71	&0.44	&0.56	&0.31	&0.34	&0.45	&0.35	\\
					&GAIN	&0.66	&0.45	&0.38	&0.02	&0.01	&0.2	&\textbf{0.71}	&0.71	&0.42	&0.54	&0.34	&\textbf{0.5}	&\textbf{0.64}	&0.43	\\
					&MDIOT	&0.62	&0.38	&0.35	&0.03	&0.02	&0.19	&\textbf{0.71}	&0.71	&0.38	&\textbf{0.57}	&0.34	&\textbf{0.5}	&0.63	&0.42	\\
					&VAEAC	&\textbf{0.77}	&\textbf{0.47}	&0.41	&0.03	&0.03	&0.19	&0.55	&0.65	&\textbf{0.65}	&0.53	&0.33	&\textbf{0.5}	&\textbf{0.64}	&0.44	\\
					&KPOD	&0.59	&0.18	&0.48	&0.07	&0.35	&0.19	&0.56	&0.67	&0.38	&0.38	&0.33	&0.44	&0.55	&0.4	\\
					&TDM	&0.57	&0.38	&0.34	&0.03	&0.02	&0.18	&0.59	&0.71	&0.4	&\textbf{0.57}	&0.34	&0.47	&0.59	&0.4	\\
					&HyperImpute	&0.74	&\textbf{0.47}	&0.4	&0.02	&0.03	&0.15	&0.58	&0.67	&0.64	&0.46	&0.31	&\textbf{0.5}	&0.63	&0.43	\\
					&MBFS	&0.74	&0.46	&0.43	&0.03	&0.03	&0.22	&0.45	&0.53	&\textbf{0.65}	&0.39	&0.25	&\textbf{0.5}	&-\tnote{1}	&0.36	\\
					&DIMV	&0.67	&0.36	&0.39	&0.03	&0.02	&0.2	&0.59	&0.71	&0.57	&\textbf{0.57}	&0.34	&0.4	&0.48	&0.41	\\
					
					\bottomrule
					
					\toprule
					\multicolumn{2}{c}{Dataset} & \emph{Iris} & \emph{Brea} & \emph{Wine} & \emph{Bank} & \emph{Htru} & \emph{Know} & \emph{Bir1} & \emph{Bir2} & \emph{Ove1} & \emph{Ove2} & \emph{Worm} & \emph{Urba} & \emph{Winn} & \textbf{Avg}\\
					\midrule
					\multirow{11}{*}{PUR}
					&\textbf{SDC (Ours)}	&0.87	&\textbf{0.9}	&\textbf{0.87}	&\textbf{0.8}	&\textbf{0.96}	&\textbf{0.69}	&0.37	&\textbf{0.65}	&0.92	&\textbf{0.6}	&\textbf{0.25}	&0.54	&0.79	&\textbf{0.71}	\\
					&MICE	&0.88	&0.85	&0.69	&0.62	&0.91	&0.52	&0.29	&0.38	&\textbf{0.93}	&0.54	&0.21	&\textbf{0.69}	&0.82	&0.64	\\
					&LRC	&0.65	&0.83	&0.62	&0.59	&0.91	&0.46	&0.29	&0.38	&0.83	&0.57	&0.21	&0.54	&0.7	&0.58	\\
					&GAIN	&0.85	&0.85	&0.66	&0.59	&0.91	&0.52	&\textbf{0.39}	&0.39	&0.81	&0.55	&0.21	&\textbf{0.69}	&0.86	&0.64	\\
					&MDIOT	&0.85	&0.82	&0.64	&0.61	&0.91	&0.52	&\textbf{0.39}	&0.39	&0.78	&0.57	&0.21	&\textbf{0.69}	&0.86	&0.63	\\
					&VAEAC	&\textbf{0.91}	&0.86	&0.69	&0.61	&0.91	&0.54	&0.3	&0.37	&\textbf{0.93}	&0.55	&0.21	&0.68	&0.86	&0.65	\\
					&KPOD	&0.77	&0.71	&0.69	&0.62	&0.94	&0.51	&0.27	&0.34	&0.77	&0.32	&0.22	&0.6	&0.71	&0.57	\\
					&TDM	&0.82	&0.82	&0.64	&0.61	&0.91	&0.52	&0.29	&0.39	&0.78	&0.57	&0.21	&0.64	&0.82	&0.62	\\
					&HyperImpute	&0.88	&0.86	&0.7	&0.59	&0.91	&0.5	&0.3	&0.4	&\textbf{0.93}	&0.54	&0.22	&\textbf{0.69}	&\textbf{0.87}	&0.65	\\
					&MBFS	&0.89	&0.85	&0.72	&0.6	&0.91	&0.53	&0.34	&0.35	&\textbf{0.93}	&0.57	&0.21	&0.68	&-\tnote{1}	&0.58	\\
					&DIMV	&0.85	&0.81	&0.68	&0.61	&0.91	&0.52	&0.3	&0.39	&0.89	&0.56	&0.21	&0.55	&0.68	&0.61	\\
					
					\bottomrule
				\end{tabular}
			\end{sc}
		\end{small}
		\begin{tablenotes}
			\footnotesize
			\item[1] Runtime is too long to get results.
		\end{tablenotes}
	\end{center}
	\end{threeparttable}
	\vskip -0.1in
\end{table*}
\end{document}